\documentclass[runningheads]{llncs}

\usepackage{eccv}

\usepackage{eccvabbrv}

\usepackage{graphicx}
\usepackage{amsmath}
\usepackage{booktabs}
\usepackage{epsfig}
\usepackage{tabularx}
\usepackage{verbatim}            
\usepackage{relsize}
\usepackage{multirow}
\usepackage{scrextend}           
\usepackage{array}               
\usepackage{makecell}            
\usepackage{subcaption}          
\usepackage[nointegrals]{wasysym}
\usepackage{lipsum}              
\usepackage{esvect}              
\usepackage{soul}                
\usepackage{float}
\usepackage{caption}
\usepackage{bm}
\usepackage{enumitem}            
\usepackage{empheq}
\usepackage{gensymb}             
\usepackage[thicklines]{cancel}  
\usepackage{arydshln}            
\usepackage{pifont}              

\setlist[itemize, 1]{label =\raisebox{-0.05\height}{\scalebox{1.4}{\textbullet}}}
\setlist[itemize]{noitemsep, topsep=0cm, leftmargin=3mm}
\allowdisplaybreaks              
\graphicspath{{figs/}}

\usepackage[accsupp]{axessibility}  

\definecolor{sns_yellow}{rgb}{1.0, 0.83, 0.59}
\definecolor{sns_violet}{rgb}{0.34, 0.08, 0.49}
\definecolor{sns_orange}{rgb}{0.78, 0.24, 0.45}
\definecolor{sns_blue}{rgb}{0.2, 0.06, 0.40}
\definecolor{cvprblue}{rgb}{0.21,0.49,0.74}
\colorlet{my_gray}{gray!10}
\definecolor{darkGreen}{rgb}{0.01, 0.8, 0.24}

\newcommand{\noIndentHeading}[1]{\noindent\textbf{#1}}

\newcommand{\cmark}{\checkmark}
\newcommand{\xmark}{\ding{53}}

\newcommand{\scaleFraction}{.85}

\newcommand{\myTopRule}{\Xhline{2\arrayrulewidth}}

\newcolumntype{t}{!{\vrule width 1.5\arrayrulewidth}}
\newcolumntype{m}{!{\vrule width 2.5\arrayrulewidth}}

\providecommand\rightarrowRHD{\relbar\joinrel\mathrel\RHD}

\newcommand{\uparrowRHD}  {\rotatebox[origin=c]{90}{$\rightarrowRHD$}}
\newcommand{\downarrowRHD}{\rotatebox[origin=c]{270}{$\rightarrowRHD$}}
\newcommand{\uparrowRHDSmall}  {\raisebox{0.05\normalbaselineskip}{\scalebox{0.7}{\uparrowRHD}}}   
\newcommand{\downarrowRHDSmall}{\raisebox{0.07\normalbaselineskip}{\scalebox{0.7}{\downarrowRHD}}} 

\newcommand{\twoD}{$2$D\xspace}
\newcommand{\threeD}{$3$D\xspace}

\newcommand{\kitti}{KITTI\xspace}
\newcommand{\nuscenes}{nuScenes\xspace}
\newcommand{\waymo}{Waymo\xspace}

\newcommand{\val}{Val\xspace}

\newcommand{\first}[1]{$\textcolor{sns_blue}{\mathbf{#1}}$}
\newcommand{\second}[1]{$\textcolor{sns_orange}{\mathbf{#1}}$}
\newcommand{\firstKey}[1]{\textcolor{sns_blue}{\textbf{#1}}}
\newcommand{\secondKey}[1]{\textcolor{sns_orange}{\textbf{#1}}}
\newcommand{\sota}{SoTA\xspace}

\newcommand{\fcosThreeD}{FCOS3D\xspace}

\newcommand{\nerf}{NeRF\xspace}
\newcommand{\nerfs}{NeRFs\xspace}
\newcommand{\autoRF}{AutoRF\xspace}
\newcommand{\autoRFWithFCOS}{\autoRF{}\!+\!FCOS\xspace}
\newcommand{\bootInv}{BootInv\xspace}
\newcommand{\codeNerf}{CodeNeRF\xspace}

\newcommand{\supNeRF}{SUP-NeRF\xspace}
\newcommand{\supBoot}{SUP-BootInv\xspace}

\newcommand{\methodName}{SUP-NeRF\xspace}

\usepackage[pagebackref,breaklinks,colorlinks,citecolor=eccvblue]{hyperref}

\usepackage{orcidlink}

\usepackage[capitalize]{cleveref}
\crefname{section}{Sec.}{Secs.}
\Crefname{section}{Section}{Sections}
\Crefname{table}{Table}{Tables}
\crefname{table}{Tab.}{Tabs.}

\newcommand{\paperTitle}{SUP-NeRF: A Streamlined Unification of Pose Estimation and NeRF for Monocular 3D Object Reconstruction}

\begin{document}
\title{\paperTitle} 

\titlerunning{\methodName}


\author{Yuliang Guo$^{1}$\thanks{Project Lead}\qquad~ Abhinav Kumar$^{1,2}$\qquad~ Cheng Zhao$^{1}$\\
Ruoyu Wang$^{1}$\qquad~~~~~~Xinyu Huang$^{1}$\qquad~~~~~~~Liu Ren$^{1}$~~\vspace{-0.2cm}}
\institute{
{$^{1}$Bosch Research North America \& Bosch Center for Artificial Intelligence (BCAI)\\
$^{2}$Michigan State University}\\
{\small\tt$^{1}$[yuliang.guo2,cheng.zhao,ruoyu.wang,xingyu.huang,liu.ren]@us.bosch.com}\\
{\small\tt$^{2}$kumarab6@msu.edu}\\
{\small\url{https://yuliangguo.github.io/supnerf}}
}

\authorrunning{Y.~Guo et al.}

\maketitle
\vspace{-0.9cm}
\begin{figure}[H]
    \centering
    \includegraphics[width=0.87\textwidth]{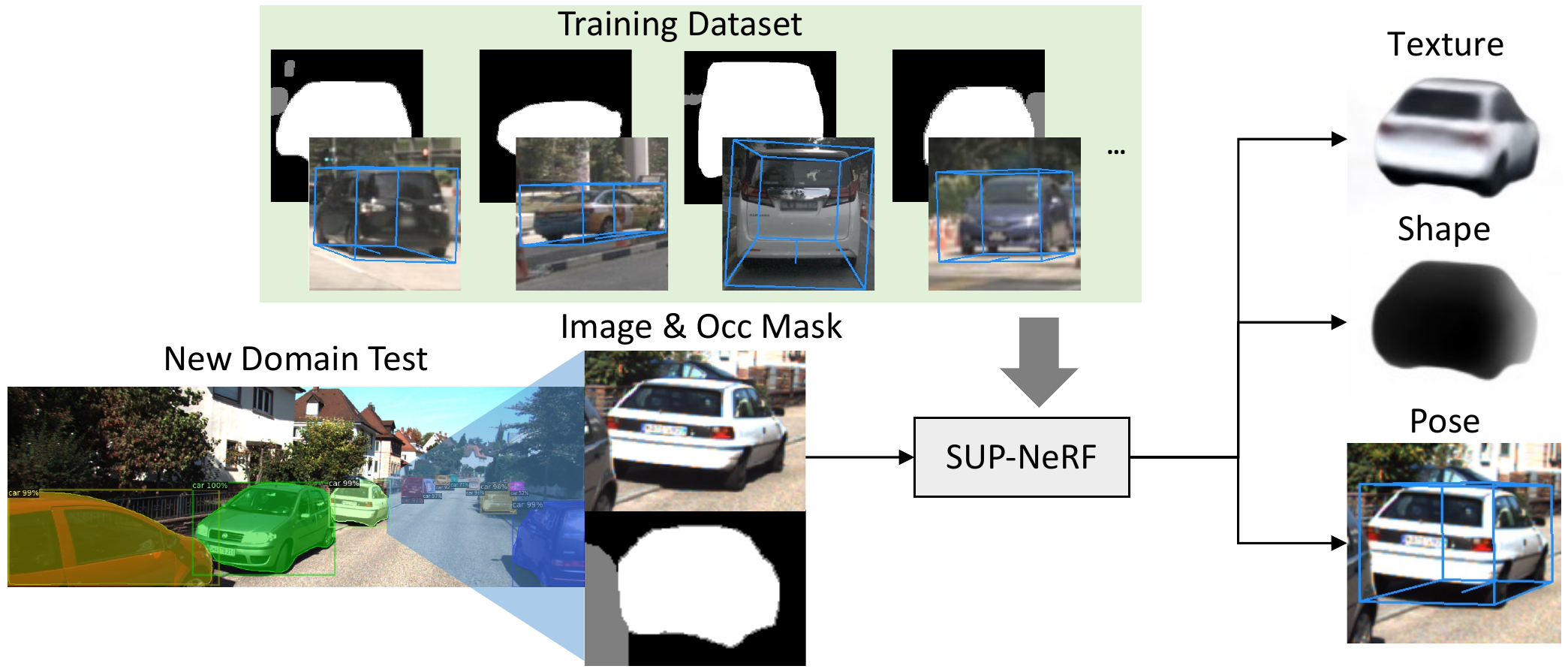}
    \vspace{-0.2cm}
    \caption{\textbf{Teaser.} \methodName is a unified solution that predicts an object's pose, shape, and texture using a single network. 
    \methodName is trained on real driving scenes with imprecise labels, and it adapts robustly to new cross-dataset scenarios.
    }
    \label{fig:teaser}
    \vspace{-0.7cm}
\end{figure}

\begin{abstract}
Monocular 3D reconstruction for categorical objects heavily relies on accurately perceiving each object's pose. 
While gradient-based optimization in a NeRF framework updates the initial pose, this paper highlights that scale-depth ambiguity in monocular object reconstruction causes failures when the initial pose deviates moderately from the true pose.
Consequently, existing methods often depend on a third-party 3D object to provide an initial object pose, leading to increased complexity and generalization issues. 
To address these challenges, we present SUP-NeRF, a Streamlined Unification of object Pose estimation and NeRF-based object reconstruction. 
SUP-NeRF decouples the object's dimension estimation and pose refinement to resolve the scale-depth ambiguity, and introduces a camera-invariant projected-box representation that generalizes cross different domains. While using a dedicated pose estimator that smoothly integrates into an object-centric NeRF, SUP-NeRF is free from external 3D detectors.
SUP-NeRF achieves state-of-the-art results in both reconstruction and pose estimation tasks on the nuScenes dataset. 
Furthermore, SUP-NeRF exhibits exceptional cross-dataset generalization on the KITTI and Waymo datasets, surpassing prior methods with up to 50\% reduction in rotation and translation error.
\end{abstract}

\section{Introduction}

Monocular \threeD object reconstruction~\cite{conf/cvpr/HendersonTL20/2d2tex3d,iccv/Gkioxari0M19/meshrcnn,conf/eccv/KanazawaTEM18/catmeshfromimage,conf/cvpr/KunduLR18/3drcnn,cvpr/ZakharovKBG20:autolabeling} is a critical technology with broad applications in autonomous driving~\cite{cvpr/ZakharovKBG20:autolabeling,conf/cvpr/KunduGYFPGTDF22/pnf,muller2022autorf}, AR/VR, robotics~\cite{RoomR}, and embodied AI~\cite{Zhou_2023_CVPR,Tian_2023_CVPR,Batra:etal:corr:2020:rearrangement}. 
Its impact is particularly significant in autonomous driving, where it plays a vital role in tasks like auto-labeling~\cite{cvpr/ZakharovKBG20:autolabeling} and world-model construction~\cite{conf/cvpr/KunduGYFPGTDF22/pnf,muller2022autorf}. This technology is essential for reconstructing 3D models from typical driving scenarios, where single-view observations are common. 
This paper focuses explicitly on the problem of recovering an object's pose, shape, and texture from a single image under real driving scenarios, see \cref{fig:teaser}. Our aim is to devise an efficient solution, thereby unlocking the full potential of monocular reconstruction for active tasks such as planning and control.

Neural Radiance Field (\nerf) technology~\cite{mildenhall2021nerf,corl/ZakharovAGPKDTS21,DBLP:journals/tog/MullerESK22:instantnerf} has revolutionized the field of \threeD reconstruction. 
It excels at capturing fine details in a scene and generating novel views from existing reconstructions. 
An exciting branch, Object-centric \nerf \cite{iccv/Yang0XLZB0C21:objectnerf,cvpr/YuYTK21:pixelnerf,conf/iccv/JangA21/codenerf,muller2022autorf,conf/cvpr/KunduGYFPGTDF22/pnf,conf/eccv/InsafutdinovCHV22/symnerf,Pavllo_2023_CVPR}, takes things further, allowing for flexible data creation and tackling reconstruction challenges from limited number of views by incorporating category-specific prior knowledge.
Earlier object-centric \nerf methods~\cite{iccv/Gkioxari0M19/meshrcnn,conf/eccv/KanazawaTEM18/catmeshfromimage,conf/cvpr/KunduLR18/3drcnn,corl/ZakharovAGPKDTS21,cvpr/ZakharovKBG20:autolabeling,cvpr/YuYTK21:pixelnerf,conf/iccv/JangA21/codenerf} require densely overlapping views and precise poses, limiting their use to synthetic data. Later advances~\cite{muller2022autorf,conf/cvpr/KunduGYFPGTDF22/pnf} enable monocular training with imperfect masks. 

\begin{figure}[!tb]
    \centering
    \includegraphics[width=0.7\textwidth]{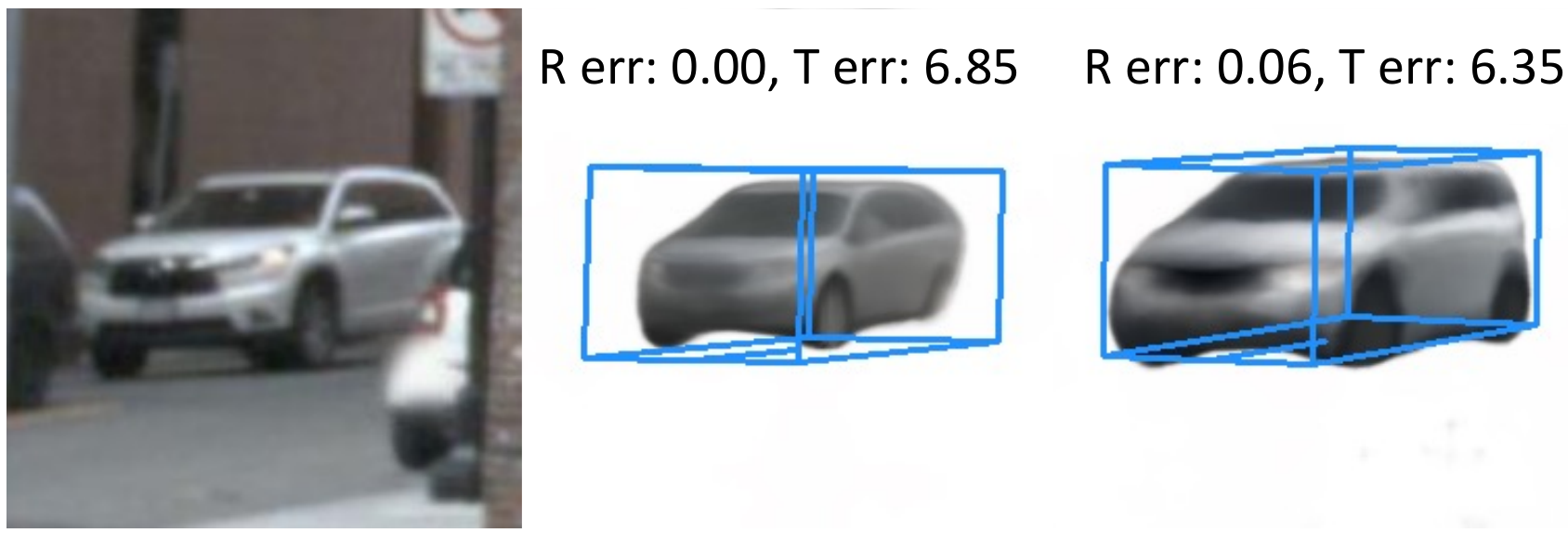}
    \caption{\textbf{Scale-depth ambiguity in \nerf}. Given the input image (left), joint optimization of pose, shape, and texture in \nerf has full freedom to rescale the shape within the normalized shape space (blue box) or move the \threeD box. Such phenomenon is observed from the evolution of the rendered objects from iteration 0 (middle) to iteration 50 (right).
    }
   \label{fig:illus:ambiguity}
   \vspace{-0.6cm}
\end{figure}

Although promising, object \nerfs show limitations in correcting erroneous object poses.
First, the well-known scale-depth ambiguity~\cite{iccv/SimonelliBPLK19:disentanglemono3d,iccv/LuMYZL0YO21:GUP,eccv/KumarBCPL22:deviant} also affects the object \nerf optimization. 
\cref{fig:illus:ambiguity} shows the optimization failure of a monocular \nerf method under longitudinal translation error when it is given complete freedom in adjusting the scale within the normalized space. 
Secondly, \nerf also struggles with large rotation errors. 
\cref{sec:GBP:impact} shows that \nerf typically requires the initial rotation error to be below $25^\circ$, and the pose optimization cannot go lower than $7^\circ$.

Recent papers~\cite{conf/cvpr/KunduGYFPGTDF22/pnf,muller2022autorf} use an external monocular \threeD detector to provide initial poses to \nerf. However, monocular \threeD detectors suffer from generalization issues \cite{eccv/KumarBCPL22:deviant,kumar2024seabird} and increased complexity, ultimately affecting the \nerf pipeline. \bootInv \cite{Pavllo_2023_CVPR}, a more recent method, tackles pose estimation, shape, and appearance reconstruction in a single framework but requires additional pre-training of dense field generator and struggles with handling moderate occlusions.

To resolve the scale-depth ambiguity and achieve greater generalization, this paper proposes \methodName, a Streamlined Unification of Pose estimation and \nerf. \methodName uses an image encoder to output the object dimensions and then refines for object pose using an iterative process while keeping its size fixed. 
Besides decoupling of an object's dimension and pose to resolve the scale-depth ambiguity, the introduction of a novel projected-box representation in the iterative update refinement also enhances the generalization to novel data since the representation makes the deep pose refiner independent from camera intrinsic parameters.
In addition, \methodName tackles the complexity challenge by sharing the \nerf image encoder with the pose estimation, making the pose estimation module a seamless fit for object \nerfs.

In our experiments, we first comprehensively benchmark and conduct ablations on \methodName on the \nuscenes \cite{caesar2020nuscenes} dataset. 
\methodName outperforms state-of-the-art (\sota) methods and achieves \sota object reconstruction and pose performance on this dataset. 
We next conduct cross-dataset object reconstruction and pose estimation experiments of the \nuscenes models on the \kitti and \waymo datasets to test \methodName's generalization capability. 
\methodName shows more significant improvements over \sota in such cross-dataset settings.

In summary, our main contributions include:
\begin{itemize}
    \item We introduce \methodName, an streamlined method unifying object pose estimation and monocular object-centric neural reconstruction.
    \item We identify depth-scale ambiguities as the key barrier to object \nerf's pose optimization and craft a module to overcome this effectively, while also integrating seamlessly with existing object \nerf frameworks.
    \item We introduce a novel projected box representation for the pose estimation module, enabling the deep refiner to operate in a space invariant to camera intrinsic parameters, thereby enhancing \methodName's generalization capability.
    \item \methodName achieves \sota object reconstruction and pose estimation results, outperforming other methods on the \nuscenes dataset, and in challenging cross-dataset experiments on both \kitti and \waymo datasets.
\end{itemize}

\section{Related Work}

\noIndentHeading{Object-centric \nerf.}
Object-centric \nerf~\cite{cvpr/YuYTK21:pixelnerf,conf/iccv/JangA21/codenerf,muller2022autorf,iccv/Yang0XLZB0C21:objectnerf,conf/cvpr/KunduGYFPGTDF22/pnf,conf/eccv/InsafutdinovCHV22/symnerf,Pavllo_2023_CVPR} is a specialized \nerf variant for modeling specific object categories. 
Its pipeline feeds a \nerf decoder with shape and texture codes, generating occupancy and color values at 3D positions, followed by volumetric rendering to produce a rendered image. 
The entire process is differentiable, allowing back-propagation of the photometric loss to update shape codes, texture codes, network weights, and object poses. 
Early object-centric \nerfs~\cite{iccv/Gkioxari0M19/meshrcnn,conf/eccv/KanazawaTEM18/catmeshfromimage,conf/cvpr/KunduLR18/3drcnn,corl/ZakharovAGPKDTS21,cvpr/ZakharovKBG20:autolabeling,cvpr/YuYTK21:pixelnerf,conf/iccv/JangA21/codenerf,iccv/Yang0XLZB0C21:objectnerf} required accurate object poses, full visibility, and dense multi-view observations to reconstruct a categorical model, making them inaccessible for real-world applications. 
Some approaches~\cite{conf/iros/LinFBRIL21:inerf,conf/iccv/JangA21/codenerf} claim successful object pose recovery through gradient-based pose refinement from monocular images.
However these successes are still limited to synthetic dataset, and oversimplified object/camera poses. 
Recent methods~\cite{conf/cvpr/KunduGYFPGTDF22/pnf,muller2022autorf} investigate more practical pipelines demonstrating the object-\nerf's potential to train and test on driving scenes with limited views and occluded object data. 
However, both methods require relatively accurate poses from heavy third-party \threeD detectors~\cite{iccv/SimonelliBPLK19:disentanglemono3d,DBLP:conf/iccv/ParkAG0G21}.
Other recent works~\cite{Bian_etal_CVPR_2023_NoPeNeRF,Chen_etal_CVPR_2023_DBARF} try freeing \nerf from external camera poses; however, solving camera poses from densely overlapping views is a different problem than recovering an object's pose from a single image.
Compared to these methods, \methodName recovers object pose from a single image without a \threeD detector in the pipeline. A more recent work~\cite{Pavllo_2023_CVPR} also unifies pose estimation and NeRF in a single network, and enables high-fidelity object generation. However, its training requires additional NOCS dense generator and it is highly sensitive to moderate occlusions compared to our method.

\begin{table}[!tb]
\caption{\textbf{Literature Summary}. 
\methodName sets a benchmark for minimal requirements, eliminating the need for external 3D detectors, CAD models, or dense NOCS generators. It is dedicated to the reconstruction of object shapes and pose estimation directly from a single image of driving scenes, employing an object-centric framework.
} 
\label{tab:liter:review}
\centering
\scalebox{0.85}{
\setlength\tabcolsep{0.08cm}\begin{tabular}{l m c|c|c|c|c|c|c|c}
\myTopRule
\multicolumn{1}{l m}{Methods}                                              & Indoor       & Outdoor       & Views             & Scope           & CAD            & wt. Det3D          & wt. \nerf              & NOCS Gen. \\
\myTopRule
6DoF Pose \cite{conf/eccv/LiWJXF18/deepim,cvpr/MerrillGZ0LPR022:suoslam}   &   \cmark     &    \xmark     &    single         &   object        &   \cmark       &     \cmark         &     \xmark         &     \xmark \\
3D Detector\cite{iccvw/WangZPL21:fcos3d,eccv/KumarBCPL22:deviant}          &   \xmark     &    \cmark     &    single         &   image         &   \xmark       &     \cmark         &     \xmark         &     \xmark \\
\nerf \cite{mildenhall2021nerf}                                            &   \cmark     &    \cmark     &    multi          &   scene         &   \xmark       &     \xmark         &     \cmark         &     \xmark \\
Object-\nerf \cite{muller2022autorf,conf/cvpr/KunduGYFPGTDF22/pnf}         &   \xmark     &    \cmark     &    single         &   object        &   \xmark       &     \xmark         &     \cmark         &     \xmark \\
\bootInv \cite{Pavllo_2023_CVPR}                                           &   \xmark     &    \cmark     &    single         &   object        &   \xmark       &     \cmark         &     \cmark         &     \cmark \\
\rowcolor{my_gray}\textbf{\supNeRF}                                         &   \xmark     &    \cmark     &    single         &   object        &   \xmark       &     \cmark         &     \cmark         &     \xmark \\
\myTopRule
\end{tabular}
}
\vspace{-0.8cm}
\end{table}

\noIndentHeading{Monocular 3D Object Pose Estimation.}
Advancements in deep learning and the availability of \threeD-labeled datasets~\cite{caesar2020nuscenes,cvpr/GeigerLU12:kitti,Hodan2018ECCV:BOP} enable many monocular \threeD object detection \cite{Park2019ICCV,kumar2021groomed,iccv/LuMYZL0YO21:GUP,eccv/KumarBCPL22:deviant,iccv/SimonelliBPLK19:disentanglemono3d,cvpr/Yu0Y22,cvpr/LipsonTG022:CoupledIter} and 6DoF pose estimation methods~\cite{conf/iccv/KehlMTIN17:ssd6d,cvpr/TekinSF18:seamless6d,cvpr/MousavianAFK17:zoox,Li2018ECCV,Xiang2018RSS,Zakharov2019ICCV,conf/eccv/LiWJXF18/deepim,cvpr/MerrillGZ0LPR022:suoslam}. 
Although estimating \threeD information from a single image is ill-posed due to depth-scale ambiguity, leveraging constraints like CAD models or sizes enables accurate object pose recovery via PnP processes~\cite{Rad2017ICCV:bb8,cvpr/TekinSF18:seamless6d,cvpr/Wang0HVSG19:nocs} or iterative optimization~\cite{conf/iccv/KehlMTIN17:ssd6d,conf/eccv/LiWJXF18/deepim}. These indoor 6DoF pose estimation mostly follow an object-centric design which aims to estimate highly accurate object pose within the \twoD image patches out of a previous \twoD object detector. Recent studies~\cite{Chen_etal_CVPR_2023_TexPose,Li_etal_corr_2023_NeRFPose} use the \nerf pipeline to reconstruct \threeD models or to train model-based keypoint detectors to estimate object poses during inference. However, none of these model-based methods is ideal for outdoor driving scenarios, as pre-reconstruction of models is often impossible in new scenes.
Fortunately, inferring the \threeD dimensions of objects from an image is feasible when its appearance is tightly bounded to its fine-grained class~\cite{cvpr/MousavianAFK17:zoox}, which keeps an outdoor \threeD detector from being fully ill-posed. Nowadays, reliable outdoor \threeD detectors usually favors a image-centric design to direct detect many 3D objects captured within the whole image, which usually requires a large backbone network~\cite{iccvw/WangZPL21:fcos3d,eccv/KumarBCPL22:deviant}. However, the performance of \sota detectors drops significantly when applied to datasets with distribution different from the training data~\cite{iccvw/WangZPL21:fcos3d,iccv/LuMYZL0YO21:GUP,eccv/KumarBCPL22:deviant}. 

\cref{tab:liter:review} outline the differences in methodologies. 
Unlike existing approaches, \methodName unifies pose estimation and object \nerf in a streamlined manner, adopting a straightforward object-centric design akin to indoor techniques. 
Without the need for CAD models, \methodName is suitable for outdoor driving scenarios. Despite its focused scope on object-centric pose estimation, \methodName showcases enhanced cross-domain generalization capabilities compared to more complex image-centric 3D detectors within this realm.



\section{\methodName}\label{sec:method}

\begin{figure}[!tb]
    \centering
    \includegraphics[width=0.98\textwidth]{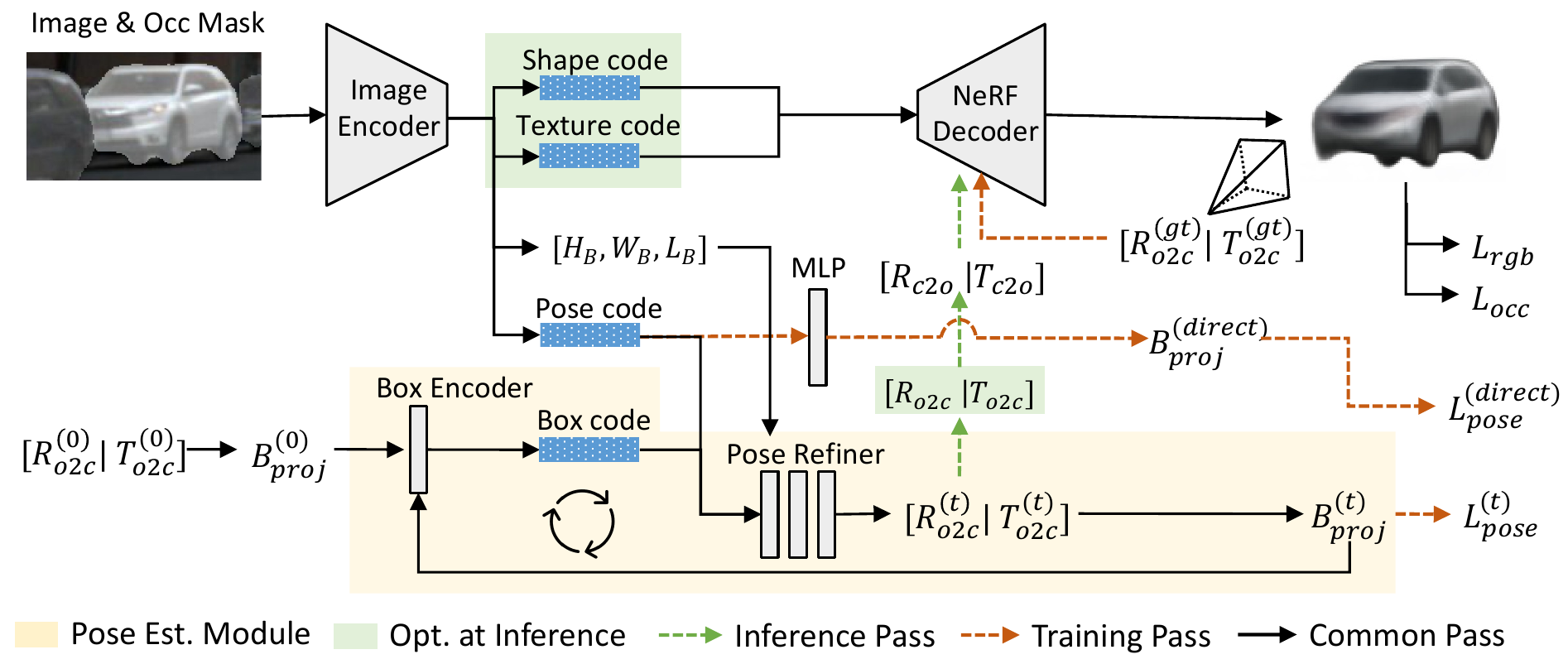}
    \caption{\textbf{\methodName Overview.} \methodName unifies pose estimation and \nerf. The pose estimation module enables \methodName to work for objects in diverse poses \textbf{without} external \threeD detectors. The increase of complexity only constitutes a few MLP layers.
    }
  \label{fig:overview}
   \vspace{-0.4cm}
\end{figure}

Our goal is to create a \nerf framework that bypasses \nerf's limitations without external 3D detectors. \cref{fig:overview} shows the overall pipeline of \methodName, a streamlined unified network that predicts object pose, shape, and texture from a single image.
\methodName consists of three modules: a Resnet-based encoder, a pose estimation module, and a \nerf decoder.  \methodName processes an image and its occlusion mask, with the encoder turning the masked image into shape, texture, and pose codes and predicted object dimensions. The pose code, object dimensions and a box code, are passed to the pose estimation module to iteratively refine the object pose $[R_{o2c} |T_{o2c}]$. After a few iterations, the estimated pose is converted to the camera pose $[R_{c2o} |T_{c2o}]$ and passed to the \nerf decoder. In the \nerf phase, the decoder uses the shape, texture codes, and camera pose for volumetric rendering, creating both RGB and occupancy images. 


\begin{figure}[!t]
    \centering
    \includegraphics[width=0.7\textwidth]{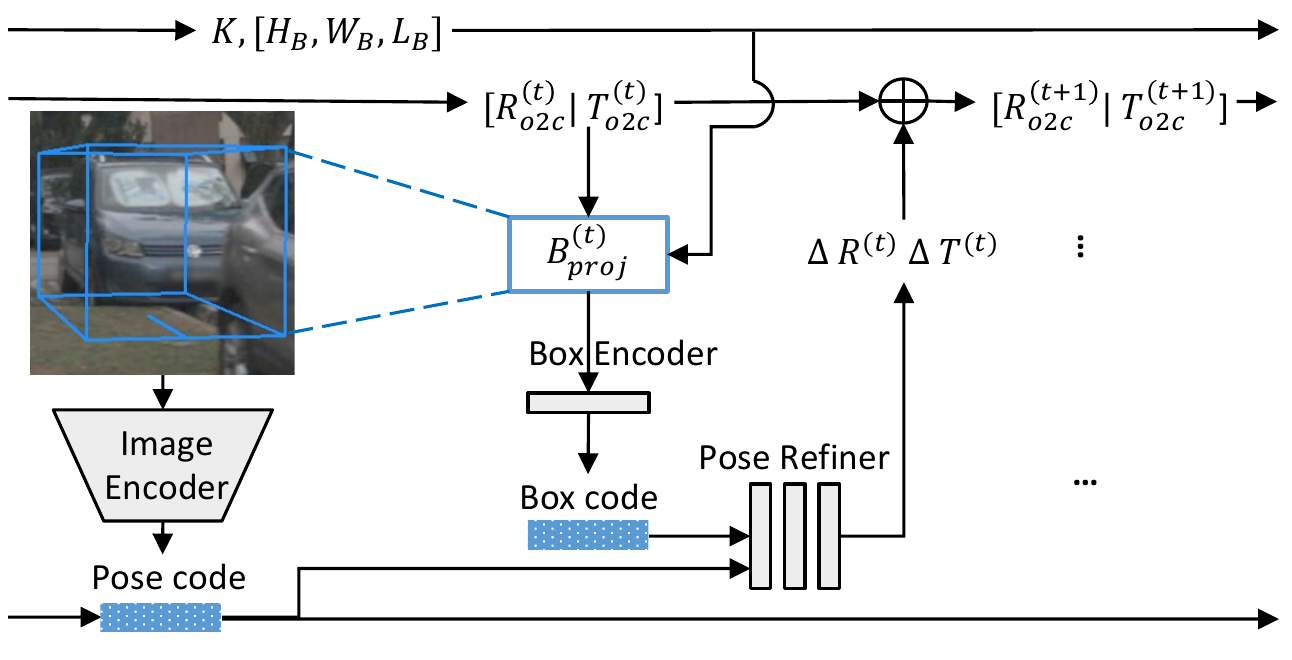}
    \caption{
    \textbf{\methodName Pose Estimation Module}. The pose estimation module of \methodName iteratively updates the object's pose while preserving scale. 
    It takes the projection of \threeD box corners as a visual representation of the input pose and estimate the pose update via comparing it to observed image in a latent embedding space. These designs handle scale-depth ambiguity and make the deep refiner independent from camera intrinsic parameters for better cross-domain generalization.
    }
  \label{fig:illus:pose:rep}
   \vspace{-0.4cm}
\end{figure}

\noIndentHeading{Pose Estimation Module.} 
The key component of \methodName is the pose estimation module, which we show in \cref{fig:illus:pose:rep}. 
It aims to provide reliable object poses for targets at all ranges and orientations.
Our pose estimation module iterative updates the input pose, based on the visual difference between the input pose and the observed object in the image. Given predicted object dimensions $[H_B, W_B, L_B]$, camera intrinsic $K$, and current pose $R^{(t)}, T^{(t)}$, image projections of the \threeD box corners $B^{(t)}_{proj}$ is computed, which can be interpolated as a visual representation of current pose. 
After encoding the $8$ corners into a higher dimensional box code, \methodName feeds into the pose refiner along with a pose code from the image encoder to predict the desired pose update $\Delta R^{(t)}, \Delta T^{(t)}$. 
\methodName then composes the pose changes with current pose to compute the next pose state, $R^{(t+1)}, T^{(t+1)}$. 
This updating process starts from a fully random object pose within the viewing frustum of the image patch, and takes a few iterations such that the final pose approaches the true pose even from a random initial pose.

The pose estimation module incorporates the following unique designs:
\begin{itemize}
    \item \methodName preserves the object dimensions to ensure unambiguous pose updates. 
    Our implementation, using projected \threeD boxes on the image plane, leads to a concise encoding of shape and pose, reducing computation in image rendering and encoding per iteration, as opposed to previous methods~\cite{conf/eccv/LiWJXF18/deepim,muller2022autorf} relying on full CAD models. Since this representation makes the deep pose refiner's task independent from camera intrinsic parameters, it leads to better generalization to different testing domains, as evidenced in \cref{tab:exp:main:kitti,tab:exp:main:waymo}.
    \item \methodName updates the object pose iteratively rather than conduct a direct pose estimation using a PnP-based~\cite{conf/iccv/KehlMTIN17:ssd6d} because the later is sensitive to missed or swapped corners thus often produces far out-of-distribution poses. Our ablation studies in \cref{tab:ablation:pose:estimator} confirm this difference. 
    \item \methodName represents the rotation update in axis-angle representation~\cite{journals/corr/abs-1812-01537/microlietheory}. 
    During rotation updating process, we convert the axis-angle space 3D rotation $\Delta R^{(t)}$ back to $SO(3)$, compose it with the previous rotation to compute the new rotation state as $R^{(t+1)}_{o2c}$. 
    \item \methodName represents the translation update in a relative space, indicated as $\Delta T^{(t)}:[v^{(t)}_x, v^{(t)}_y, \rho^{(t)}]$, where $v^{(t)}_x$ and $v^{(t)}_y$ indicate the location change in image space, and $\rho^{(t)}$ indicates the ratio between the target depth and the source depth. 
    Specifically, \methodName updates the translation in image space and depth as:
    \vspace{-0.5cm}
    \begin{align}\label{eqn:trans:update}
            x^{(t+1)} &= x^{(t)} + v^{(t)}_x \nonumber\\
            y^{(t+1)} &= y^{(t)} + v^{(t)}_y \\
            Z^{(t+1)} &= \rho^{(t)} Z^{(t)} \nonumber
    \end{align}
\end{itemize}
where $(x, y)$ represent the object's \twoD location in image space, $Z$ indicates the depth. The new \threeD location $T^{(t+1)}_{o2c}$ is then computed from the updated image coordinates, depth and camera intrinsic matrix $K$. Compared to~\cite{conf/eccv/LiWJXF18/deepim,Pavllo_2023_CVPR}, our definition of $\rho^{(t)}$ leads to a more straight-forward update. 

During the pose updating process indicated by $\oplus$, both the rotation and translation are updated. Through such representation and updating scheme, \methodName handles objects at varying distances with the same output space. Further discusses the choice of an effective pose representation and the right coordinate frame are included in \cref{sec:pose:rep,sec:coord:choice}.

\noIndentHeading{Cross-task Shortcut.} 
Simultaneously perceiving pose, shape, and texture, requires an image encoder with shared initial layers but separate later stages. 
However, encoding shape, texture, and pose is a challenge as they require different features, making it tough for a shared encoder to meet all needs effectively. 
Shape and texture encoding demand a pose-invariant approach, as they correspond to each object's frame. 
On the other hand, pose encoding needs to capture sufficient pose information for later use. 
This difference leads to conflicting learning signals during joint training, potentially weakening the model's performance.

\methodName proposes a straightforward one-step approach to resolve this conflict. 
After processing through the first four CNN layers separately, \methodName introduces a cross-task shortcut that deducts pose features from the shape and texture features. 
This design choice eliminates the shape and texture’s reliance on pose, allowing the earlier layers to focus on pose-dependent features for all tasks. This shortcut turns the conflict into a cooperative relationship, enhancing the joint training of different modules, as demonstrated in \cref{tab:ablation:model:training}. The details of our image encoder is described in Appendix~\ref{sec:encoder}.

\noIndentHeading{Training.}
\methodName closely follows the loss functions and occupancy mask definitions from \autoRF \cite{muller2022autorf}. The total loss during training is: 
$L_{train} = L_{rgb} + w_{occ} L_{occ} + w_{pose} (L^{(direct)}_{pose} + \sum^K_t L^{(t)}_{pose})$, 
where $w_{occ}$ and $w_{pose}$ indicate the corresponding weights. 

We compare the RGB output to the input image to calculate photometric loss  $L_{rgb}$, and the occupancy values to the mask for occupancy loss $L_{occ}$. 
\methodName additionally compares the projected box corners from the integrative pose estimator ${B^{(t)}}_{proj}$ with the ground-truth values to compute the pose losses ${L^{(t)}_{pose}}$. We define the pose loss function as the mean distance between the matched corners: 
$L^{(*)}_{pose} = \dfrac{1}{8} \sum\limits^{8}_{i=1} \sqrt{(x^{(*)}_i - \hat{x}_i)^2 + (y^{(*)}_i - \hat{y}_i)^2}$,
where $(*)$ indicates it compatible with both the direct pose loss and the iterative pose loss, $(x^{(*)}_i, y^{(*)}_i)$ denote a input corner position, and $(\hat{x}_i, \hat{y}_i)$ indicates a ground-truth corner position.
\methodName introduces an additional MLP layer for more accurate pose estimation to directly supervise the \twoD coordinates of box corners predicted from the pose code. 

In training, only the network weights are treated as the optimizable variables. Moreover, the \nerf decoder uses ground-truth camera poses $[R^{gt}_{c2o} |T^{gt}_{c2o}]$ instead of predictions from the pose estimation module to ensure more reliable training.

\noIndentHeading{Implementation Details.}
\methodName's image encoder backbone is based on ResNet50, with shared early layers and separated later layers for pose, shape, and texture encoding. For more information on this, please refer to the \cref{sec:encoder}. To implement the \nerf decoder, we use \codeNerf~\cite{conf/iccv/JangA21/codenerf} for its simplicity and ease of use. For the differentiable rendering based on \nerf, we set the latent dimension to 256, use 64 samples along each ray, and normalize the shape space by the diagonal length of the observed object's \threeD box. For the pose estimation module, we also set the latent dimension to 256 for all MLP layers. Before feeding the input image and occupancy mask into the image encoder, we pad them to make them square and resize them to $128\times128$ pixels. We then render $32\times32$ images and occupancy maps to calculate losses.

To ensure effective iterative pose estimation without sacrificing generalization, \methodName uses a heuristic to sample a random initial \threeD object pose across the entire pose space. For translation, we sample in the relative space. We sample a random \twoD location within the image ROI range as the initial projection of the object center, while setting the initial depth to $Z=20$. For rotation, we sample from the full range of yaw angles combined with a random rotation angle within a $(-20^\circ, 20^\circ)$ range around a random axis direction. This sampling method covers almost the entire distribution of object poses in the camera frame and is used consistently in both training and inference.

During training, we use 3 iterations of the pose estimator to generate corresponding pose losses, which allow for pose regression to handle highly diverse poses. We \textit{jointly} train the unified model for the pose estimation module, image encoder, and \nerf decoder for 40 epochs on the \nuscenes dataset using a learning rate of $10^{-4}$ and loss weights of $w_{occ} = 0.1$, $w_{poss} = 0.01$ to compute the total loss. 

\noIndentHeading{Inference.} During inference, \methodName optimizes the shape code, texture code, and object pose $[R_{o2c} |T_{o2c}]$ keeping the network weights frozen.
The total loss $L_{infer}$ is thus $L_{infer} = L_{rgb} + w_{occ} L_{occ}$, where $w_{occ}$ balances the two. 
\methodName minimizes the loss $L_{infer}$ to update the variables.
At inference, we run 3 iterations for the \textit{feed-forward} pose estimation module without updating the shape and color codes, and 50 iterations for the \textit{gradient-based} joint optimization of shape, texture, and pose. We use a step size of $0.02$ for shape and texture code updates and $0.01$ for pose updates during joint optimization.
\section{Experiments}
\vspace{-0.4cm}
Our experiments use three datasets: \nuscenes \cite{caesar2020nuscenes}, \kitti \cite{cvpr/GeigerLU12:kitti} and \waymo \cite{sun2020scalability}. 
Our evaluation framework simultaneously measures rendering quality, shape reconstruction accuracy, and pose estimation precision. 

\noIndentHeading{Data Splits.}
Since these datasets are not specifically for \threeD reconstruction, we sifted a subset for better evaluation. 
We chose daytime sequences and used Mask R-CNN~\cite{Detectron2018} to get instance masks, as \nuscenes lacks \twoD segmentation masks. 
We matched these masks with \threeD bounding box annotations, categorizing them into foreground, background, and unknown, similar to \autoRF \cite{muller2022autorf}. More details are in \cref{sec:data:prepare}. 
We use the following splits of these datasets:
\begin{itemize}
    \item \textit{\nuscenes Train Split}: We collected $87{,}048$ objects from the \nuscenes Train split for training. 
    \item \textit{\nuscenes \val Split}: We collected $5{,}000$ random objects from the \nuscenes \val split for evaluation. 
    \item \textit{\kitti \val Split}: We use $4{,}895$ objects from \kitti \val split \cite{cvpr/GeigerLU12:kitti} for evaluation.
    \item \textit{\waymo \val Split}:  We randomly selected $5{,}000$ objects from \waymo \val split \cite{eccv/KumarBCPL22:deviant} for evaluation.
\end{itemize}

\begin{figure}[!t]
    \centering
    \includegraphics[width=0.95\textwidth]{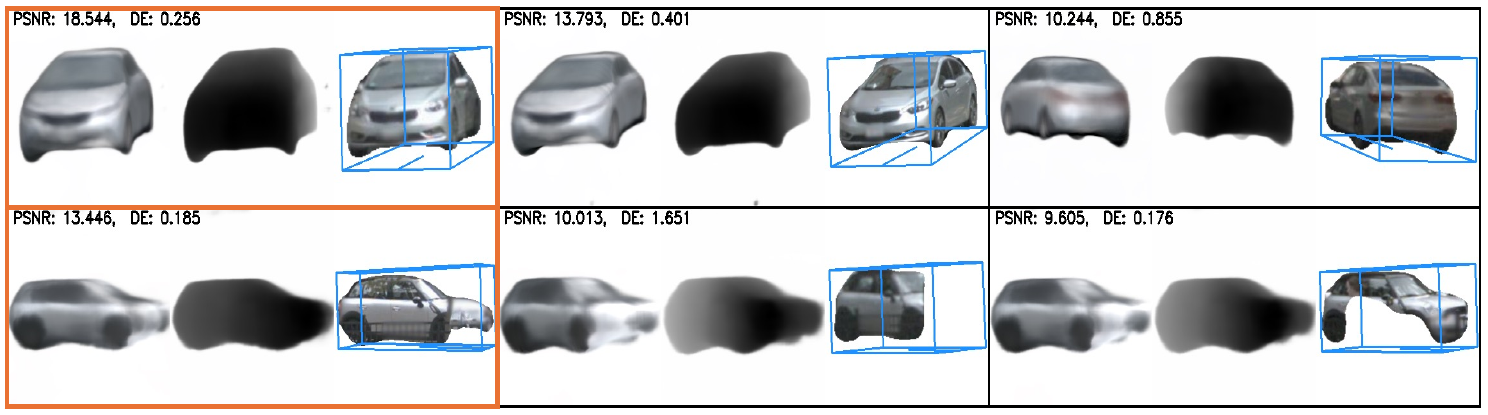}
    \caption{\textbf{\nuscenes Cross-view Evaluation}. 
    Each row shows a set of images of the same object from different angles. We use the \textcolor{orange}{example} for both monocular pose estimation and \nerf reconstruction. Other images in each row are used to evaluate the reconstructed shape and texture in PSNR and Depth Error (DE).}
  \label{fig:exp:cross:eval}
\end{figure}

\noIndentHeading{Evaluation metrics.} Our evaluation framework differs from traditional \nerf assessments as it simultaneously evaluates object reconstruction quality and pose accuracy. We incorporate two monocular reconstruction metrics:
(1) the \textit{PSNR} (Peak Signal-to-Noise Ratio) comparing the rendered and observed images;
(2) the Depth Error between the rendered depth map and sparse LiDAR data, denoted as \textit{DE}.
We also include two pose estimation metrics: 
(3) rotation error denoted as \textit{RE}; (4) the translation error denoted as \textit{TE}. 
To offer a thorough evaluation, we compare various methods at different stages: post-\textit{feed-forward (FF)}, and at the 50th iterations of \nerf \textit{joint optimization}.
Beyond monocular assessments, we also conduct \textit{cross-view} evaluations for PSNR and depth error metrics, denoted as (5) \textit{PSNR-C} and (6) \textit{DE-C}, focusing on objects observed from novel views. As the \nuscenes dataset provides object IDs that allows tracking objects across various views, the cross-view evaluation is exclusively conducted on the \nuscenes dataset, as explained in \cref{fig:exp:cross:eval}. 

\noIndentHeading{Baselines.} 
We compare \methodName with two baselines: \autoRF \cite{muller2022autorf} and \autoRF\cite{muller2022autorf}+\fcosThreeD \cite{iccvw/WangZPL21:fcos3d}(selected for its monocular capability, and availability of model trained on the \nuscenes dataset).
Both \autoRF and \supNeRF use the same \nerf decoder and image encoder to ensure a fair comparison. The difference lies in \supNeRF's additional layers for pose encoding and refinement. 
During testing, we apply the same random pose generation for both \autoRF and \supNeRF methods, whereas the \fcosThreeD baseline begins with its own pose prediction. 
We also compare with \bootInv \cite{Pavllo_2023_CVPR}, which recovers pose, shape, and appearance. 
Leveraging our flexible design, we developed \supBoot by integrating \bootInv's pretrained decoder with our encoder and pose estimator.

\begin{table}[!tb]
\caption{\textbf{\nuscenes Monocular Reconstruction and Pose Estimation Results}. 
\supNeRF consistently improves AutoRF-based pipelines in all metrics, particularly in the pose estimation metrics. \supBoot also consistently improves \bootInv in all metrics.
[Key: \firstKey{Best}, \secondKey{Second Best}, FF= Feed Forward, C= Cross-View]
} 
\label{tab:exp:main:nusc}
\centering
\begin{footnotesize}
\scalebox{\scaleFraction}{
\setlength\tabcolsep{0.08cm}
\begin{tabular}{lmc|c|c|c|c|c}
\myTopRule
\multirow{2}{*}{Method}                                 & PSNR (\uparrowRHDSmall)       & DE (m) (\downarrowRHDSmall)      & RE (deg.) (\downarrowRHDSmall)    & TE (m) (\downarrowRHDSmall)  & PSNR-C (\uparrowRHDSmall)    &DE-C (m) (\downarrowRHDSmall)        \\ \cline{2-7} 
                                                        & FF|50it                       & FF|50it                          & FF|50it                           & FF|50it                      & FF|50it                      & FF|50it                             \\ \myTopRule
\autoRF\cite{muller2022autorf}                          & 3.6|10.6                      & 11.21|10.09                      & 87.52|88.07                       & 6.04|5.95                    & 10.0|8.8                     & 1.31|1.41                           \\ 
\autoRFWithFCOS                                         & 7.5|\second{17.2}             & \second{1.34|0.81}               & 9.77|10.17                        & 0.85|\second{0.78}           & 9.8|10.5                     & \second{1.29|1.16}                  \\ 
\rowcolor{my_gray}\textbf{\supNeRF}                      & \second{10.5}|\first{18.8}    & \first{0.69|0.60}                & \first{7.01|7.07}                 & \second{0.68}|\first{0.73}   & \second{10.6}|10.9           & \first{1.22|1.14}              \\ \hline

\bootInv\cite{Pavllo_2023_CVPR}                         & 9.4|14.3                      & 5.01|3.56                        & 28.40|28.00                       & 2.59|2.91                    & \first{10.9}|\second{11.8}   & 1.37|1.35                           \\ 
\rowcolor{my_gray}\textbf{\supBoot}                      & \first{10.9}|15.4             & 1.95|1.62                        & \second{7.11|8.40}                & \first{0.64}|1.00            & \first{10.9|11.9}            & 1.37|1.40                           \\ \myTopRule
\end{tabular}
}
\end{footnotesize}
\end{table}

\subsection{\nuscenes Monocular Reconstruction and Pose Estimation}\label{sec:nusc:main}

\cref{tab:exp:main:nusc} shows that \supNeRF outperforms two baselines, \autoRF and its combination with \fcosThreeD on \nuscenes \val set. 
\autoRF alone struggles to improve pose from a random start, resulting in lower PSNR and depth error. 
\supNeRF consistently reduces depth error by over 25\%, rotation error by over 30\%, and translation error by over 13\% compared to the \fcosThreeD and \autoRF combination. 
Our feed-forward stage also shows promising potential for real-time applications, especially where rendering quality is less critical. 

We next compare \bootInv \cite{Pavllo_2023_CVPR} as the baseline, although the performance of original \bootInv's pose estimation was subpar on both the \nuscenes and \kitti datasets, integrating our robust pose module with this \nerf framework significantly enhanced pose estimation accuracy. 
This underscores the adaptability of our unified pipeline in augmenting various object \nerf frameworks for improved pose estimation. 
Moreover, comparisons of \bootInv \cite{Pavllo_2023_CVPR} and \supBoot in terms of monocular PSNR and depth error suggest that \supBoot's superior pose estimation contributes to its higher PSNR and lower depth error. More detailed analysis upon \bootInv-based pipelines is included in \cref{sec:main:exp:ext}.

\begin{table}[!tb]
\caption{\textbf{\kitti Cross-dataset Monocular Reconstruction and Pose Estimation Results}. 
We train all methods on \nuscenes dataset, and test on \kitti dataset. Our methods \supNeRF and \supBoot consistently show superiority in all metrics compared to the counterpart methods.
[Key: \firstKey{Best}, \secondKey{Second Best}, FF= Feed Forward]
}
\label{tab:exp:main:kitti}
\centering
\begin{footnotesize}
\scalebox{\scaleFraction}{
\setlength\tabcolsep{0.15cm}
\begin{tabular}{lmc|c|c|c}
\myTopRule
\multirow{2}{*}{Method}                 & PSNR (\uparrowRHDSmall)                & DE (m) (\downarrowRHDSmall)         & RE (deg.) (\downarrowRHDSmall)    & TE (m) (\downarrowRHDSmall)   \\ \cline{2-5} 
                                        & FF|50it                                & FF|50it                             & FF|50it                           & FF|50it                       \\ \myTopRule
\autoRF\cite{muller2022autorf}          & 0.4|7.6                                & 9.89|8.83                           & 89.67|90.42                       & 6.16|6.07                      \\ 
\autoRFWithFCOS                         & 2.4|\second{13.4}                      & \second{2.42|1.74}                  & 11.95|12.5                       & 2.2|2.09                      \\
\rowcolor{my_gray}\textbf{\supNeRF}    & 4.0|\first{14.1}                       & \first{2.19|1.54}                   & \first{6.79|6.89}                 & \first{1.06|1.01}              \\ \hline
\bootInv\cite{Pavllo_2023_CVPR}         & \second{7.0}|12.3                      & 6.11|4.38                           & 15.52|15.76                       & 3.97|3.90                      \\ 
\rowcolor{my_gray}\textbf{\supBoot}      & \first{7.6}|12.7              & 2.79| 2.21                          & \second{9.16|8.84}                & \second{1.15|1.67}             \\ \myTopRule
\end{tabular}
}
\end{footnotesize}
\end{table}

\begin{figure*}[!t]
    \centering
    \includegraphics[width=0.95\textwidth]{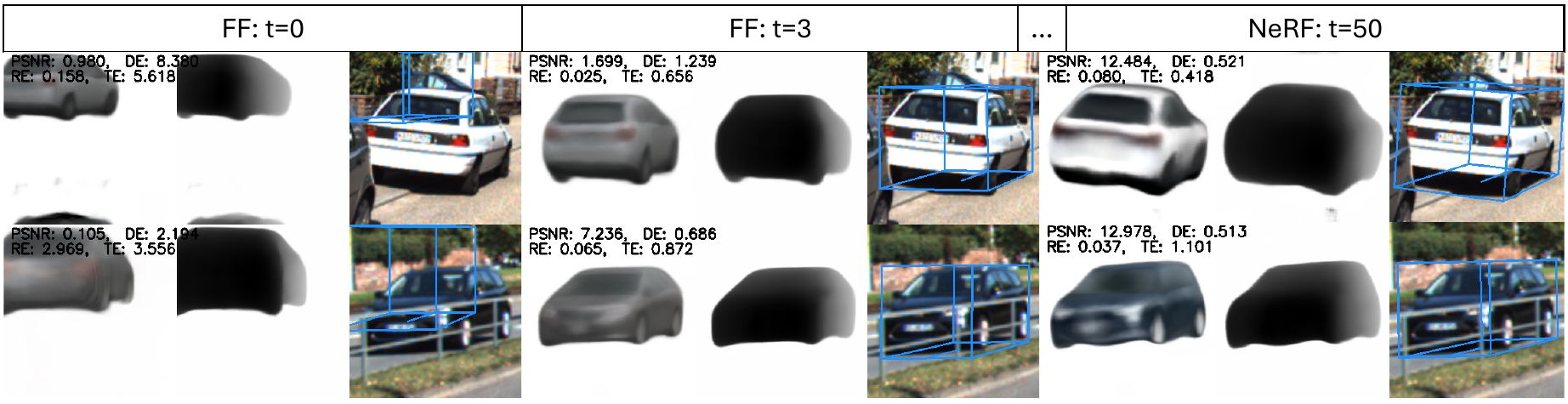}
    \includegraphics[width=0.95\textwidth]{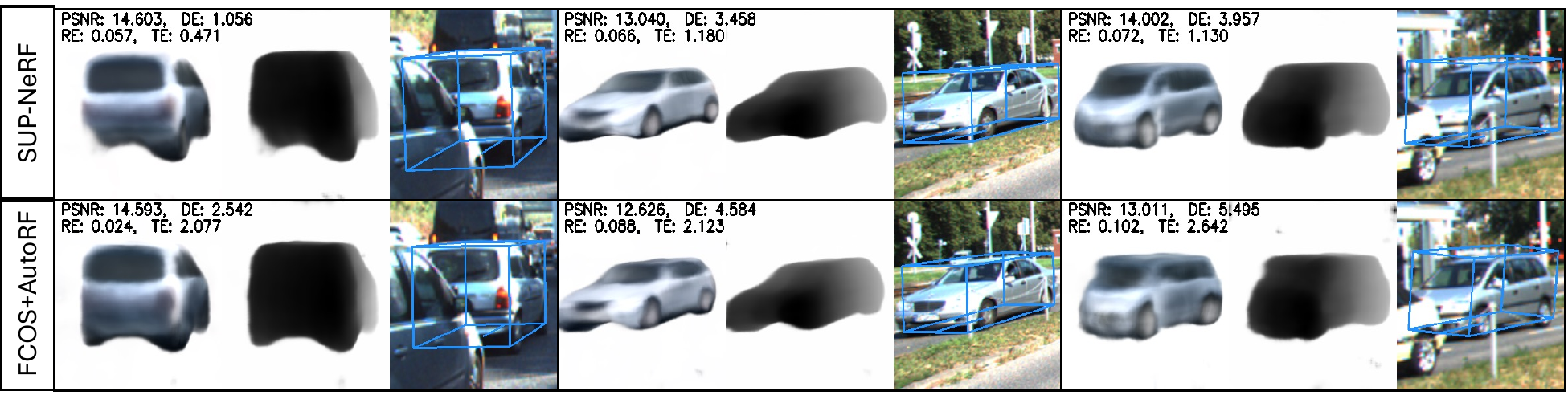}
    \caption{\textbf{\kitti (Cross-dataset) Qualitative Results}. In the top panel, we demonstrate \supNeRF executes pose estimation reliably, fast converging from a random initial pose to the true one, and enables neural reconstruction under diverse object poses, occlusion cases under this cross-dataset setup. In the bottom panel, \supNeRF is visually compared to the other major competitor, demonstrates sharper rendered image, higher accuracy in shape and pose.
    }
  \label{fig:exp:visual:results}
\end{figure*}

\subsection{Cross-Dataset Evaluation}
We next compare the generalization capability of all these \nerf models with the cross-dataset experiments. For these experiments, we directly apply \nuscenes models to the \kitti and \waymo datasets and report the metrics. 

\noIndentHeading{\kitti.} \cref{tab:exp:main:kitti} shows the result of cross-dataset experiments on the \kitti dataset. 
It shows that \supNeRF achieves \sota performance across all evaluation metrics. 
\supNeRF reduces depth error by about 16\%, rotation error by as large as 50\% and translation error also about 50\%. 
Thus, \supNeRF shows significantly better geometric performance in cross-dataset setting compared to \autoRFWithFCOS, which we attributes to our deep pose module's independence from camera intrinsic parameters.
Meanwhile, compared to \nuscenes evaluation, all the methods' PSNR and depth error performance drop, which means the cross-dataset generalization of monocular \nerf remains an open challenge. 

\cref{fig:exp:visual:results} presents the qualitative results on this experiment, demonstrating how \supNeRF handles diverse poses and occlusions in pose refinement and the visual comparisons between the two most competitive pipelines. Consistent improvements upon integrating our design into \bootInv baseline were also observed. More visual comparisons are in \cref{sec:more:vis}.

\noIndentHeading{\waymo.} \cref{tab:exp:main:waymo} shows the result of cross-dataset experiments on the \waymo dataset. 
It shows that \supNeRF achieves \sota performance across most evaluation metrics. The only exception is the superior rotation estimation achieved by \fcosThreeD, which is likely due to leveraging larger context for heavily occluded cases. However, our advantage in translation estimation is more significant.

\begin{table}[!tb]
\caption{\textbf{\waymo Cross-dataset Monocular Reconstruction and Pose Estimation Results}. 
We train all methods on \nuscenes dataset, and test on \waymo dataset. Our methods \supNeRF and \supBoot consistently show superiority in all metrics compared to the counterpart methods, the only exception is the comparison to AutoRF+FCOS in Rotation Error.
[Key: \firstKey{Best}, \secondKey{Second Best}, FF= Feed Forward]
}
\label{tab:exp:main:waymo}
\centering
\begin{footnotesize}
\scalebox{\scaleFraction}{
\setlength\tabcolsep{0.15cm}
\begin{tabular}{lmc|c|c|c}
\myTopRule
\multirow{2}{*}{Method}                 & PSNR (\uparrowRHDSmall)    & DE (m) (\downarrowRHDSmall)         & RE (deg.) (\downarrowRHDSmall)    & TE (m) (\downarrowRHDSmall)   \\ \cline{2-5} 
                                        & FF|50it                    & FF|50it                             & FF|50it                           & FF|50it                        \\ \myTopRule
\autoRF\cite{muller2022autorf}          & 0.6|9.8                    & 6.76|6.53                           & 86.56|87.67                       & 9.1|9.14                       \\ 
\autoRFWithFCOS                         & 6.2|16.5                   & \second{2.43|2.32}                  & \first{6.65|7.2}                  & 3.22|3.31                      \\
\rowcolor{my_gray}\textbf{\supNeRF}    & 4.8|\first{17.0}           & \first{2.32|1.56}                   & \second{10.01|10.6}               & \second{1.68}|\first{1.54}     \\ \hline
\bootInv\cite{Pavllo_2023_CVPR}         & \second{8.1}|11.0          & 8.85|8.20                           & 30.78|31.52                       & 5.26|6.16                      \\ 
\rowcolor{my_gray}\textbf{\supBoot}      & \first{8.6}|11.9           & 5.36|4.38                           & 10.24|11.04                       & \first{1.67}|\second{2.53}     \\ \myTopRule
\end{tabular}
}
\end{footnotesize}
\end{table}

\subsection{Ablation Studies}

The major ablation studies we conduct includes (i) the impact of pose estimation module designs and (ii) the impact of different object \nerf networks. Other ablation studies to analyze the impact of training with predicted \twoD boxes, pose refinement iterations, training epochs, as well as \nerf's capability on handling initial pose errors are included in \cref{sec:ablation:more}, owing to lack of space.

\noIndentHeading{Pose Estimation Module.} 
We evaluate multiple pose estimation designs from two main angles. First, we focus on object-centric pose estimation task alone with evaluations on \nuscenes, \kitti, and \waymo datasets. The prototypical designs include: (i) Direct 6D pose regression using MLP layers; (ii) Pose estimation from predicted box corners via Keypoint-based Perspective-n-Point (PnP); (iii) NOCS-based pose estimation as implemented in \bootInv~\cite{Pavllo_2023_CVPR}; (vi) External monocular \threeD detection \fcosThreeD~\cite{iccvw/WangZPL21:fcos3d}; (v) Our iterative pose refinement method, which uses projected boxes. For fairness, methods (1), (2), and (5) share the same image encoder, pose encoding layers, differing only in their output layers. \cref{tab:ablation:pose:module} shows that \methodName consistently outperforms the others, notably in cross-dataset translation estimation.

The second evaluation assesses the impact of different initial poses on object-NeRF performance within the \supNeRF architecture. We additionally compared (i) Ground-Truth (GT) pose; (ii) Random initial pose, to represent the oracle and worst case in pose initialization, and exclude the NOCS method which is part of a different NeRF framework. Given different initial poses, all methods undergo the same \nerf Gradient-based Pose Refinement (NGPR), evaluated immediately and after 50 iterations. \cref{tab:ablation:pose:estimator} shows that our \methodName method significantly surpasses other initial pose estimators in geometric accuracy, demonstrating its effectiveness in enhancing 3D model quality. 

\begin{table}[!tb]
\caption{\textbf{Pose Estimation Ablation Studies.} 
We evaluate all the candidate pose estimation methods with a focus on pose accuracy alone, on \nuscenes, \kitti, and \waymo validation sets. 
Our method shows outstanding robustness in all cases, particularly in translation accuracy in cross-dataset tests.
[Key: \firstKey{Best}]
}
\label{tab:ablation:pose:module}
\centering
\scalebox{\scaleFraction}{
\begin{tabular}{lmccc|ccc}
\myTopRule
\multirow{2}{*}{Pose Module}                & \multicolumn{3}{c|}{RE (deg.)(\downarrowRHDSmall)}          & \multicolumn{3}{c}{TE (m) (\downarrowRHDSmall)}    \\ \cline{2-7} 
                                            & \nuscenes & \kitti & \waymo                                 & \nuscenes & \kitti & \waymo                \\ \myTopRule
MLP Pose                                    & 31.96 & 37.58 & 30.89                                       & 5.53 & 5.31 & 10.1                  \\ 
Corners+PnP                                 & 24.79 & 35.10 & 93.76                                       & 2.82 & 3.55 & 18.68                 \\ 
NOCS~~+PnP(BootInv)                         & 28.40 & 15.52 & 23.66                                       & 2.59 & 3.97 & 3.47                  \\
\fcosThreeD                                 & 9.77 & 11.95 & \first{6.65}                                 & 0.85 & 2.2 & 3.22                  \\
\rowcolor{my_gray}\textbf{\methodName}   & \first{7.01}& \first{6.79} & 10.01                          & \first{0.68}& \first{1.06}& \first{1.54}          \\ \myTopRule
\end{tabular}
}
\end{table}

\begin{table}[!tb]
\caption{\textbf{Ablation studies of Pose Initialization in Object NeRF} on \nuscenes. 
We compare different choices of initial poses under \supNeRF frameworks. As shown, our \methodName presents superiors effectiveness in all the metrics.
[Key: \firstKey{Best}, FF= Feed Forward, C= Cross-View]
}
\label{tab:ablation:pose:estimator}
\centering
\scalebox{\scaleFraction}{
\begin{tabular}{lmc|c|c|c|c|c}
\myTopRule
\multirow{2}{*}{Initial Pose}                   & PSNR (\uparrowRHDSmall)      & DE (m) (\downarrowRHDSmall)  & RE (deg.) (\downarrowRHDSmall)  & TE (m) (\downarrowRHDSmall)  & PSNR-C (\uparrowRHDSmall)      & DE-C (m) (\downarrowRHDSmall)            \\ \cline{2-7} 
                                        & FF|50it                      & FF|50it                      & FF|50it                         & FF|50it                      & FF|50it                        & FF|50it                                  \\
\myTopRule
Random                                  & 3.6|10.6                     & 11.21|10.09                  & 87.52|88.07                     & 6.04|5.95                    & 9.1|7.4                        & 1.42|1.82                                \\ 
MLP Pose                                & 1.2|6.9                      & 5.33|4.41                    & 31.96|24.57                     & 5.53|2.78                    & 9.8|8.5                        & 1.3|1.4                                  \\ 
Corners+PnP                             & 7.2|16.2                     & 2.18|1.83                    & 24.79|24.57                     & 2.82|2.78                    & 9.9|9.5                        & 1.3|1.3                                  \\ 
\fcosThreeD                             & 7.5|17.2                     & 1.34|0.81                     & 9.77|10.17                      & 0.85|0.78                    & 9.8||10.5                      & 1.29||1.16                        \\ 
\rowcolor{my_gray}\textbf{\methodName}         & \first{10.5|18.8}            & \first{0.69|0.60}            & \first{7.01|7.07}               & \first{0.68|0.73}            & \first{10.6|10.9}              & \first{1.22|1.14}                                \\ \hline
GT                                      & 10.0|18.8                    & 0.66|0.53                    & 0.|2.11                         & 0.|0.13                      & 10.9|11.1                      & 1.19|1.14                                \\ \hline
\myTopRule
\end{tabular}
}
\end{table}

\cref{tab:ablation:pose:estimator} also confirms the role of \nerf Gradient-Based Pose Refinement (NGPR). 
By comparing the difference between feed-forward results and the later ones, one can find NGPR merely reduces rotation error or translation error. 
On the contrary, it can even increase pose error when the initial pose error is small enough, e.g. rotation error $< 7^\circ$, and translation error $< 0.75 m$. 
On the other hand, the improvements of PSNR and depth reconstruction over time indicate that \nerf optimization mostly focuses on shape and texture optimization \wrt the visible surface regardless the pose.

\noIndentHeading{Impact of object \nerf framework.} To demonstrate \supNeRF's benefits, we compared different network configurations using the same pose refinement approach. For fairness, we used identical subnetwork modules across all configurations. These include a ResNet-based Image Encoder (E), a \nerf Decoder (D), and our Pose Estimation Module (P). \codeNerf\cite{conf/iccv/JangA21/codenerf} has just a Decoder (D), \autoRF combines an Image Encoder and Decoder (E + D), and \supNeRF uses all three (E + D + P). For \codeNerf and \autoRF, we added a separately trained Pose Network (PNet, composed of E + P) for initial pose refinement. We also tested a \supNeRF version without the cross-task Short Cut (SC) to assess its impact. \cref{tab:ablation:model:training} shows that SC consistently improves \supNeRF across all metrics on \nuscenes.

\supNeRF, with joint training, outperforms separate training setups \autoRF + PNet and \codeNerf + PNet in monocular metrics. Interestingly, while \codeNerf lags in monocular, it excels in cross-view evaluations. This implies that using mean shape codes, rather than current image encodings, might reduce overfitting to current view. 


\begin{table}[!tb]
\caption{\textbf{Object \nerf Framework Ablations} on \nuscenes. We used the same subnetwork architectures - the image encoder (E), \nerf decoder (D), and pose estimation module (P) - to create various model configurations. For \supNeRF, we also included a version without the cross-task Short Cut (SC) for comparison. 
[Key: \firstKey{Best}, FF= Feed Forward, C= Cross-View]
}
\label{tab:ablation:model:training}
\centering
\scalebox{0.81}{
\begin{tabular}{l|c|c|c|c|c|c}
\myTopRule
\multirow{2}{*}{Model Training}  & PSNR(\uparrowRHDSmall)     & DE (m)(\downarrowRHDSmall)    & RE (deg.)(\downarrowRHDSmall)  & TE (m)(\downarrowRHDSmall)   & PSNR-C(\uparrowRHDSmall)     & DE-C (m)(\downarrowRHDSmall)    \\ \cline{2-7} 
                                 & FF|50it                     & FF|50it                        & FF|50it                         & FF|50it                       & FF|50it                       & FF|50it                          \\ \myTopRule
\codeNerf(D)+PNet(E+P)           & 7.6|16.0                    & 0.87|0.69                      & 10.57|9.75                      & 0.72|0.76                     & \first{11.0|11.2}             & \first{1.01|1.04}                \\
\autoRF(E+D)+PNet(E+P)           & 9.7|17.5                    & 0.79|0.66                      & 8.26|7.87                       & 0.7|0.74                      & 5.6|10.1                      & 1.24|1.16                        \\
\supNeRF(E+D+P) w/o SC          & 8.7|18.0                    & 0.94|0.73                      & 10.12|9.22                      & 0.75|0.74                     & 10.8|10.9                     & 1.23|1.11                        \\
\rowcolor{my_gray}\textbf{\supNeRF(E+D+P)}                 & \first{10.5|18.8}           & \first{0.69|0.60}              & \first{7.01|7.07}               & \first{0.68|0.73}             & 10.6|10.9                     & 1.22|1.14                        \\ \myTopRule

\end{tabular}
}
\end{table}

\subsection{Running Speed Analysis}

\cref{tab:run:time} compares the model size and running speed on a single A5000 Graphics card. 
Detailed setup and analysis details are in \cref{sec:detail:running:speed}.
\cref{tab:run:time} shows that \supNeRF needs about half or one third as many parameters as others and is 6-8 times faster in the feed-forward (FF) stage. 
Meanwhile, a significant portion of the run time is consumed by the \nerf optimization step, whether for 20 iterations (20it) or 50 iterations (50it). However, speed limitation of NeRF can been effectively tackled by recent advances in neural rendering~\cite{MullerESK_2022_SIGGRAPH_instantNGP,Kerbl:etal:SIGGRAPH2023:GaussianSplatting}, which we consider as orthogonal to our contribution to the feed-forward stage.

\begin{table}[!tb]
\caption{\textbf{Model Size and Running Time Comparison} in feed-forward scenarios, and with 20 and 50 iterations of \nerf optimization.
\supNeRF gets the smallest running time.
[Key: \firstKey{Best}, FF= Feed Forward]
}
\label{tab:run:time}
\centering
\scalebox{\scaleFraction}{
\begin{tabular}{l m c | c | c | c }
\myTopRule
Method                                  & Params (M) (\downarrowRHDSmall)           & FF (s)   (\downarrowRHDSmall)   & FF+20it (s)  (\downarrowRHDSmall)      & FF+50it (s)    (\downarrowRHDSmall)       \\
\myTopRule
\multirow{2}{*}{\autoRFWithFCOS~\cite{muller2022autorf}}          & 91.116                                    & 0.123                           & 0.714                                  & 1.599          \\
                                        & (36.166+54.950)                           & (0.114+0.009)                   & (0.114+0.600)                          & (0.114+1.485)       \\
\rowcolor{my_gray}\textbf{\supNeRF}             & \first{49.816}                            & \first{0.018}                   & \first{0.608}                          & \first{1.493}           \\ \hline
\bootInv~\cite{Pavllo_2023_CVPR}        & 182.616                                   & 0.156                           & 3.534                                  & 8.601           \\ 
\rowcolor{my_gray}\textbf{\supBoot}               &  57.580                                   & 0.018                           & 3.396                                  & 8.463           \\ 
\myTopRule
\end{tabular}
}
\end{table}

\section{Conclusion}
This paper introduces \methodName: a unified network that seamlessly integrates pose estimation and object reconstruction.
\methodName includes a novel pose estimation module to handle scale-depth ambiguity and a new representation invariant to camera changes.
Consequently, \methodName greatly improves the robustness of object pose estimation compared to the standard \nerf baseline. 
\methodName outperforms previous methods in both monocular reconstruction and pose estimation tasks, especially in challenging cross-dataset applications.

\clearpage  
\bibliographystyle{splncs04}


\clearpage
\appendix

\renewcommand{\thesection}{A\arabic{section}}

\section*{\Large \paperTitle\\[12pt] Appendix\\[18pt]}

{For readers with a keen interest in \textbf{extensive experiments}, particularly for the detailed interpretation of the results related to \bootInv~\cite{Pavllo_2023_CVPR}, we invite you to explore \cref{sec:main:exp:ext}. 
If you are looking for more \textbf{comprehensive ablation studies}, you can find them in \cref{sec:ablation:more}. Additionally, for \textbf{further visual comparisons} with other methods and cross-view visualizations, please refer to \cref{sec:more:vis}.

\section{Pose Representation}
\label{sec:pose:rep}

Choosing the appropriate pose representation is crucial for both object pose estimation and \nerf's optimization of object pose. For translation optimization, using a \threeD representation with a single step size may not suffice for objects at varying distances. To address this, a relative space representation, inspired by~\cite{conf/eccv/LiWJXF18/deepim,Pavllo_2023_CVPR}, can be more effective. We demonstrate in \cref{sec:method} that a relative representation benefits direct regression through a neural network by allowing predictions within a more constrained space. 


In contrast, in driving scenarios where rotation is mostly simplified to 1D, the choice of rotation representation may not be as crucial. The Axis-angle representation can be a simple and effective choice to facilitate the optimization task, as demonstrated in~\cite{journals/corr/abs-1812-01537/microlietheory}. We also adopt this representation in our pose estimation design.




\section{Coordinate Frame for \nerf Optimization}
\label{sec:coord:choice}

The success of pose optimization in a object-centric \nerf framework crucially depends on the choice of the coordinate frame~\cite{Pavllo_2023_CVPR}. This section systematically discusses the right choice of coordinate system to perform the \nerf optimization for pose. Although a \nerf framework directly uses the camera pose in the object system $P_{c2o}$ for the rendering purpose, we will later show the right choice is the object pose under camera coordinate frame $P_{o2c}$.

\begin{figure}[!t]
    \centering
    \includegraphics[width=0.47\textwidth]{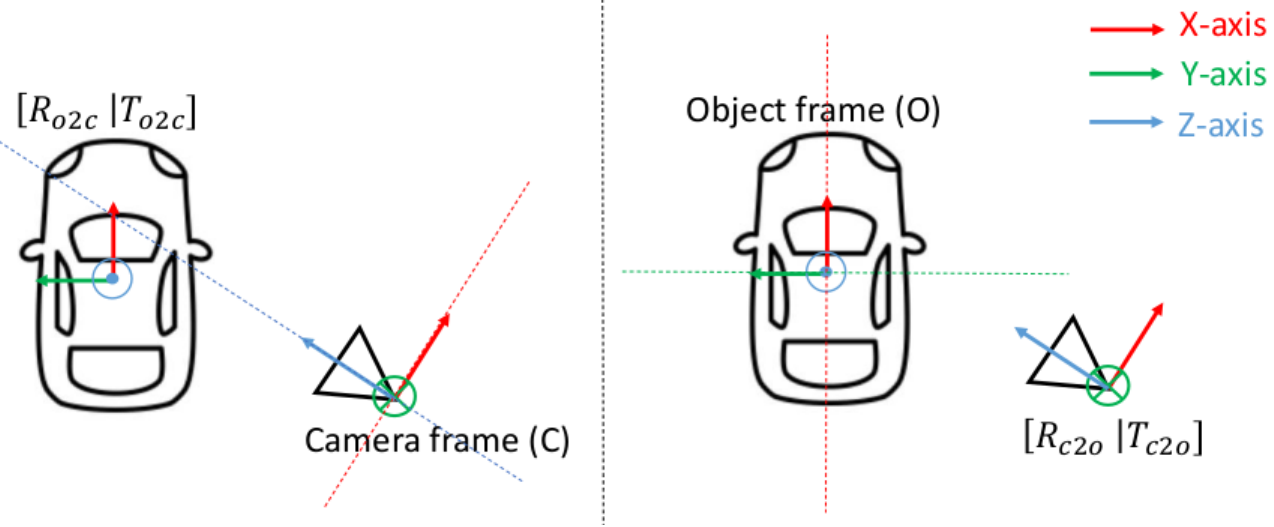}
    \caption{\textbf{Illustration of coordinate systems.} This illustration follows the \nuscenes \cite{caesar2020nuscenes} definition of camera coordinate frame (C) and object coordinate frame (O). The object orientation and location in the camera frame is denoted as $R_{o2c}, T_{o2c}$. The camera orientation and location in the object frame is denoted as $R_{c2o}, T_{c2o}$.}
  \label{fig:coord:illus}
\end{figure}

\subsection{Definitions and problem formulation}

$P_{o2c}$ and $P_{c2o}$ are defined respectively \wrt the camera frame (C) and object frame (O), as illustrated in \cref{fig:coord:illus}. $P_{o2c} = [R_{o2c} |T_{o2c}]$ indicates the transformation from object coordinates to camera coordinates, where $R_{o2c}, T_{o2c}$ indicate the object orientation (three object basis axis) and translation (object center position) in the camera frame respectively. Given $X_o, X_c$ indicating a \threeD point in the object frame and camera frame respectively, we will have the transformation
\begin{equation}
    X_c = R_{o2c} X_o + T_{o2c}
\end{equation}

Similarly, $P_{c2o} = [R_{c2o} |T_{c2o}]$ indicates the transformation from camera coordinates to object coordinates, where $R_{c2o}, T_{c2o}$ indicate the camera orientation (three camera basis axis) and translation (camera center position) in the object frame. We have
\begin{equation}
    X_o = R_{c2o} X_c + T_{c2o}
\end{equation}
We can also derive that the transformations between to two poses:
\begin{equation}
    R_{c2o} =  R_{o2c}^T ,  T_{c2o} = - R_{o2c}^T T_{o2c}
\end{equation}
and reversely 
\begin{equation}
    R_{o2c} =  R_{c2o}^T ,  T_{o2c} = - R_{c2o}^T T_{c2o}
\end{equation}

Given an object \nerf optimization process \wrt $[R_{o2c} |T_{o2c}] \rightarrow [R_{c2o} |T_{c2o}] \rightarrow L$ where $L: SE(3) \rightarrow \mathbb{R}$ denotes a loss function of $R_{c2o}, T_{c2o}$, the question can be formulated as whether the directly optimized object pose $[R_{o2c} |T_{o2c}]$ equals to the optimized camera pose $[R_{c2o} |T_{c2o}]$ transformed back to the object pose.

\subsection{Rotation representation}
\label{supp:rot:rep}

Because direct adding perturbation to the matrix might lead to a new matrix not belonging to SO(3) group, the optimization is better to b executed in another representation of rotation. An effective choice of such rotation representation can be the Lie Algebra of SO(3), the axis-angle representation~\cite{journals/corr/abs-1812-01537/microlietheory}. The transformation between the Lie Algebra of the \threeD rotation living in two coordinate frames can be written as

\begin{small}
\begin{equation}
   \begin{split}
       q  &= log(R), R = exp(q)\\
       q' &= log(R^T) = log(exp(q)^T) = log(exp(-q)) = -q
   \end{split}
\end{equation}
\end{small}
Given the axis-angle representation, the optimization can be conducted in an extended chain from an O2C pose $q_{o2c} \rightarrow [R_{o2c} |T_{o2c}] \rightarrow [R_{c2o} |T_{c2o}] \rightarrow L$ or a extended chain from a C2O pose, $q_{c2o} \rightarrow [R_{c2o} |T_{c2o}] \rightarrow L$.

\subsection{Equivalence in rotation updates?}
\label{supp:rot:equi}

The equivalence question can be answered just from the updates of rotations. Since $L$ is a function of both $R_{c2o}$ and $T_{c2o}$, each is a function of $R_{o2c}$, the updates of $R_{o2c}$ will get mixed gradients from $R_{c2o}$ and $T_{c2o}$ in separate terms. In \nerf context, given $L = G(X_o)$, where $X_o = R_{c2o} X_c + T_{c2o}$ is an \threeD point in the object frame transformed from the sampled \threeD point $X_c$ from the camera frame, the updates of $q_{o2c}$ can be written as:

\begin{small}
\begin{equation}
    \begin{split}
        {q_{o2c}^T}^{(t+1)} &= {q_{o2c}^T}^{(t)} - \lambda {\frac{\partial L}{\partial q_{o2c}}}^{(t)} \\
                            &= {q_{o2c}^T}^{(t)} - \lambda {\frac{\partial L}{\partial X_o}}^{(t)} {\frac{\partial X_o}{\partial q_{o2c}}}^{(t)} \\
    \end{split}
\label{eq:rot:update1}
\end{equation}
\end{small}
where $\lambda$ indicates the update step. Because $X_o = R_{c2o} X_c + T_{c2o}$, we have
\begin{small}
\begin{equation}
    \begin{split}
        {q_{o2c}^T}^{(t+1)} &= {q_{o2c}^T}^{(t)} - \lambda ({\frac{\partial L}{\partial X_o}}^{(t)} {\frac{\partial X_o}{\partial R_{c2o}}}^{(t)} \frac{\partial R_{c2o}}{\partial q_{o2c}}^{(t)} \\
                            &+ {\frac{\partial L}{\partial X_o}}^{(t)} {\frac{\partial X_o}{\partial T_{c2o}}}^{(t)} \frac{\partial T_{c2o}}{\partial q_{o2c}}^{(t)})\\ 
                            &= {q_{o2c}^T}^{(t)} - \lambda ({\frac{\partial L}{\partial R_{c2o}}}^{(t)} \frac{\partial R_{c2o}}{\partial q_{o2c}}^{(t)} + {\frac{\partial L}{\partial T_{c2o}}}^{(t)} \frac{\partial T_{c2o}}{\partial q_{o2c}}^{(t)})\\
                            &= {q_{o2c}^T}^{(t)} - \lambda {\frac{\partial L}{\partial q_{o2c}}}^{(t)} - \lambda {\frac{\partial L}{\partial T_{c2o}}}^{(t)} \frac{\partial T_{c2o}}{\partial q_{o2c}}^{(t)} \\
    \end{split}
\label{eq:rot:update2}
\end{equation}
\end{small}
Suppose current rotation representations in two frames are still synchronized (it might not be violated after the first iteration), that is ${q_{o2c}^T}^{(t)} = -{q_{c2o}^T}^{(t)}$, their derivatives \wrt $L$ are also each others' negative, we have 
\begin{small}
\begin{equation}
    \begin{split}
        {q_{o2c}^T}^{(t+1)} &= -({q_{c2o}^T}^{(t)} - \lambda {\frac{\partial L}{\partial q_{c2o}}}^{(t)}) - \lambda {\frac{\partial L}{\partial T_{c2o}}}^{(t)} \frac{\partial T_{c2o}}{\partial q_{o2c}}^{(t)} \\
    \end{split}
\label{eq:rot:update3}
\end{equation}
\end{small}
Since $({q_{c2o}}^{(t)} - \lambda {\frac{\partial L}{\partial q_{c2o}}}^{(t)})$ is actually the updated orientation under C2O, we can indicate it as ${q_{c2o}}^{(t+1)}$, therefore
\begin{small}
\begin{equation}
    \begin{split}
        {q_{o2c}^T}^{(t+1)} &= -{q_{c2o}^T}^{(t+1)} - \lambda {\frac{\partial L}{\partial T_{c2o}}}^{(t)} \frac{\partial T_{c2o}}{\partial R_{o2c}}^{(t)} \frac{\partial R_{o2c}}{\partial q_{o2c}}^{(t)} \\
                            &= -{q_{c2o}^T}^{(t+1)} + \lambda {\frac{\partial L}{\partial T_{c2o}}}^{(t)} \frac{\partial R_{o2c} T_{o2c}}{\partial R_{o2c}}^{(t)} \frac{\partial R_{o2c}}{\partial q_{o2c}}^{(t)} \\
    \end{split}
\label{eq:rot:update4}
\end{equation}
\end{small}
Because the existence of an additional gradient term besides $-{q_{c2o}^T}^{(t+1)}$, the optimization of $q_{o2c}$ is \textbf{not equal} to the optimization of $q_{c2o}$, which concludes the first question.

\subsection{The right choice of coordinate frame}
\label{supp:select:frame}

After understanding the optimization in the two coordinate frames are nonequivalent, the next question is about the right choice for object \nerf optimization task. One may think $P_{o2c}$ is not optimal because it is affected by a gradient entangled with both rotation and translation terms. However, in practice, which is also verified by our ablation study, \cref{sec:GBP:impact}, $P_{o2c}$ indeed performs better than $P_{c2o}$. Through our analysis, we found the devils actually in the different ranges of error to optimize within the two coordinate frames. 

As seen from \cref{fig:coord:err:illus}, if the pose error is originally generated from the camera frame, in $P_{o2c}$, the error transformed in $P_{c2o}$ can be amplified significantly because the involvement of translation term. In other words, a small pose error in the original space can be significantly amplified in translation error when the norm of the translation vector is large. This is particularly for the object \nerf framework case where an object pose $R_{o2c}, T_{o2c}$ estimated under camera frame is usually imperfect. When converting to $T_{o2c}$, due to the rotation error and far distance, the resulting error in $T_{c2o}$ can be amplified significantly. As we discussed in the main text, because optimization the translation is rather ill-posed under the joint optimization without constraining shape scale, the optimization in $P_{c2o}$ will mostly fail.

\begin{figure}[!t]
    \centering
    \includegraphics[width=0.8\textwidth]{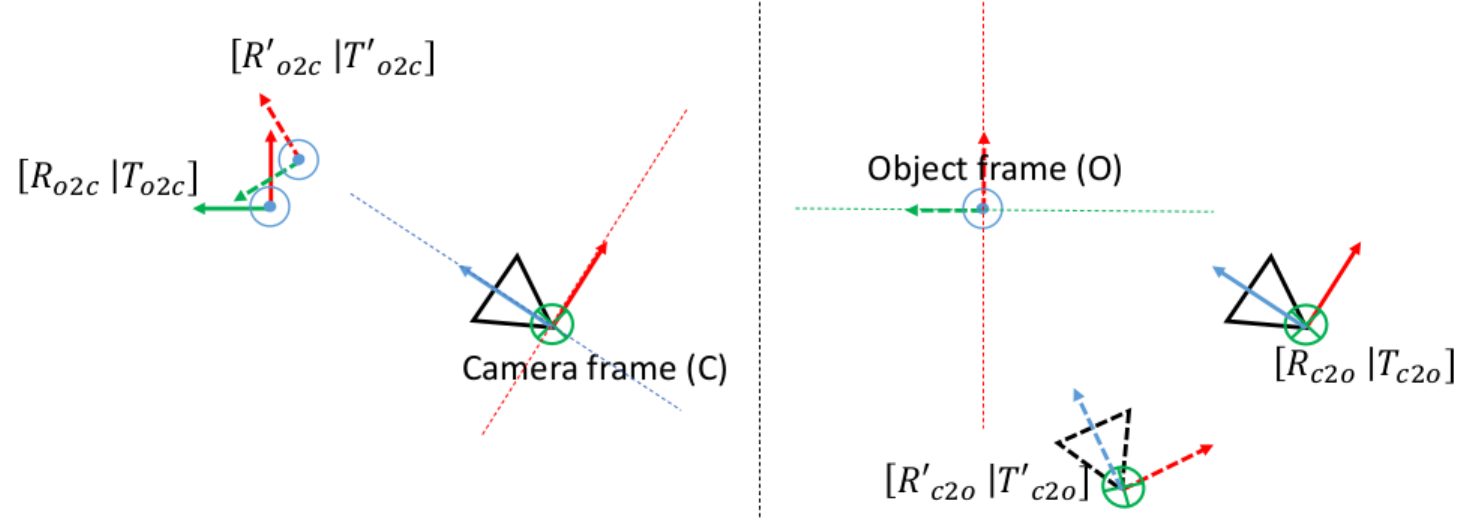}
    \caption{\textbf{Illustration of error difference under coordinate systems.} A small error in $P_{o2c}$ shown on the left can be significantly amplified in translation error in $P_{c2o}$.}
  \label{fig:coord:err:illus}
\end{figure}

This analysis tells that whether the gradient from rotation or translation are entangled is not the most critical factor, but the initial error range is. For a more concrete example, if the translation error in $P_{o2c}$ was 0.01 meter, but with some rotation error say 30 degree and the object is relatively far, the translation error transformed to $P_{c2o}$ could be over 10 meters. If a \nerf uses the later space to conduct optimization, \nerf's limitation in optimizing pose could be amplified and lead to even worse results.

\subsection{Empirical result}

We empirically confirm our analysis by comparing the two choices of coordinate frame in the \nerf pose optimization process given only a moderate rotation error but under perfect translation. As shown in \cref{fig:exp:vary:err:range}, the first two curves belong to this pair of comparison. Since the perfect translation is given, there is no much difference in depth error and translation error between the two. However, the huge advantage in the PSNR and Rotation error substantiate the right choice of $P_{o2c}$, the object pose under camera coordinate frame. Finally, we emphasis this conclusion only apply to the \textit{monocular} object pose optimization. In a typical multi-view scene reconstruction, joint optimization camera poses and reconstruction such like bundle adjustment technique is a very different setup.


\section{Encoder Details}
\label{sec:encoder}

\cref{fig:encoder:full} depicts our image encoder, which utilizes a ResNet50 backbone. The first three layers are shared across all tasks, whereas task-specific processing begins from the fourth layer onward. We introduce a feature subtraction step after the fourth layer, which separates the pose-dependent features from the shape and texture features. This separation helps to resolve conflicts between feature dependencies and enhance the synergy between pose estimation and neural reconstruction. To aid in this process, we add an additional linear layer to the pose code, which directly regresses projected corner locations during training. Another linear layer is optionally added to regress the \threeD object dimensions out of the shape code and textures code.

\begin{figure}[!t]
    \centering
    \includegraphics[width=0.9\textwidth]{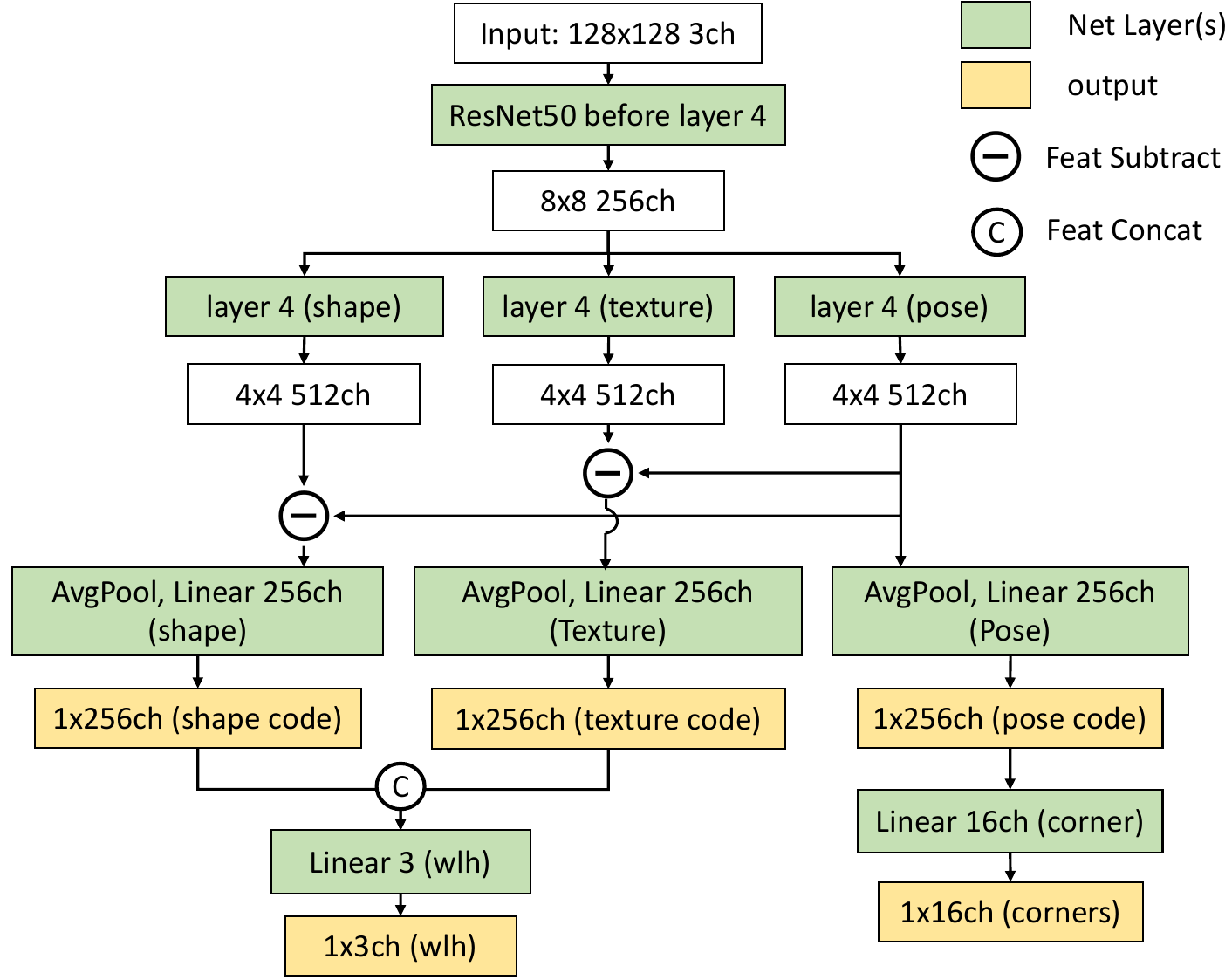}
    \caption{\textbf{Image Encoder Architecture.} Compared to the standard image encoder implemented in \autoRF, our image add layers to regress pose code and projected corners' coordinates. We also introduces cross-task feature subtraction to solve the conflict between feature's dependency to object pose.}
  \label{fig:encoder:full}
\end{figure}

\section{Outline of Additional Experiments}

In the upcoming sections, we conduct further experiments to comprehensively evaluate \methodName's capabilities and limitations, alongside those of existing approaches. In \cref{sec:main:exp:ext}, we extend our analysis of the \bootInv-related pipeline (\cite{Pavllo_2023_CVPR}) to include the \kitti dataset. We also address a \textbf{correction} in our PSNR and depth evaluations from the main paper, which initially did not adequately cater to occluded objects. \cref{sec:ablation:more} presents additional ablation studies, focusing on various configurations of \methodName and examining \nerf's constraints in pose optimization. \cref{sec:data:prepare} delves into the specifics of dataset preparation, offering detailed statistics on object perception distances across both datasets. Insights into our methodology for analyzing running speed are detailed in \cref{sec:detail:running:speed}. Finally, \cref{sec:more:vis} showcases extra visual results, further substantiating our quantitative analyses.

\section{Extended Main Experiments}
\label{sec:main:exp:ext}


In this section, we includes more detailed analysis in the comparisons with \bootInv~\cite{Pavllo_2023_CVPR}.  The integrated pipeline NPNeRF-Boot uses our pose module's output pose as the initial pose and proceeds with joint optimization of pose, shape and appearance in the same way as the original \bootInv pipeline. Moreover, the original pipeline of \bootInv requires specific adaptations to enable the evaluation on the interested datasets correctly. The major modifications are around properly using the accurate camera intrinsic parameters provided by the testing dataset, and adjusting the pre-trained model's scene range properly to the testing dataset. Besides the two pipelines, we also examined \bootInv's performance given ground-truth poses. Specifically, in the combination of ground-truth pose with \bootInv \nerf, we tested pipelines with frozen pose and jointly optimized pose separately to uncover the impact of joint optimizing of pose. Furthermore, all the candidates are additionally evaluated at the 20th iteration after gradient-based updates to better examine the trends. 

As evidenced by \cref{tab:exp:main:nusc:ext,tab:exp:main:kitti:ext,tab:exp:main:waymo:ext}, the performance of \bootInv's pose estimation was subpar on all the \nuscenes, \kitti and \waymo datasets,  However, integrating our robust pose module with this \nerf framework significantly enhanced pose estimation accuracy. This underscores the adaptability of our unified pipeline in augmenting various object \nerf frameworks for improved pose estimation. Moreover, comparisons of \bootInv and \supBoot in terms of monocular PSNR and depth error suggest that \supBoot's superior pose estimation contributes to its higher PSNR and lower depth error. 
Additionally, our extensive analysis spanning multiple \nerf frameworks and diverse initial pose settings has revealed a critical shortcoming in \nerf's gradient-based pose optimization. It has become evident that \nerf's optimization is ineffectual for enhancing poses that are either extremely poor or overly precise. This observation corroborates our earlier discussion of \nerf's limitations in pose optimization, as outlined in Section~\cref{sec:method}, and motivates the introduction of a separate pose estimation module.


In the comparison between \bootInv-based and \autoRF-based pipelines, several key observations emerge. Firstly, despite \bootInv's decoder producing highly realistic renderings of objects (as seen in \cref{fig:exp:bootinv:nusc} and \cref{fig:exp:bootinv:kitti}), its monocular PSNR often falls short of the simpler \autoRF-based pipelines. However, \bootInv-based pipelines excel in achieving higher cross-view PSNR compared to their \autoRF counterparts. This disparity stems from the distinct focuses of the two frameworks: \autoRF aims to reconstruct objects closely resembling current-view observations, leading to higher PSNR scores, whereas \bootInv incorporates more robust prior knowledge to ensure the completeness of reconstructed objects, evident in its superior cross-view PSNR. These findings highlight the ongoing research challenge of striking an optimal balance between reconstruction precision and the completeness of objects informed by prior knowledge.
Additionally, it was noted that \bootInv-based pipelines exhibit higher depth errors, pointing to a potential limitation in \bootInv's decoder in accurately perceiving scale. The difficulty in capturing the physical scales of objects with pre-trained generative models, especially in complex real-world datasets, remains a relatively uncharted area in research, underscoring the need for further exploration in this domain.

\begin{table}[ht]
\caption{\textbf{\nuscenes Monocular Reconstruction and Pose Estimation Results}. \supNeRF consistently improves AutoRF-based pipelines in all metrics, particularly in the pose estimation metrics. \supBoot also consistently improves \bootInv in all metrics. 
[Key: \textbf{Best}, FF = Feed Forward]
} 
\label{tab:exp:main:nusc:ext}
\centering
\begin{footnotesize}
\scalebox{0.72}{
\begin{tabular}{l|c|c|c|c|c|c}
\hline
\multirow{2}{*}{Method}             & PSNR (\uparrowRHDSmall)      & DE (m) (\downarrowRHDSmall)      & RE (deg.) (\downarrowRHDSmall)    & TE (m) (\downarrowRHDSmall)   & PSNR-C (\uparrowRHDSmall)    & DE-C (m) (\downarrowRHDSmall)        \\ \cline{2-7} 
                                    & FF|20it|50it                 & FF|20it|50it                     & FF|20it|50it                      & FF|20it|50it                  & FF|20it|50it                 & FF|20it|50it                         \\ \hline
GT(Frz)+\bootInv                    & 10.9|12.3|14.1               & 1.42|1.26|1.10                   & 0.00|0.00|0.00                    & 0.00|0.00|0.00                & 10.9|11.5|11.7               & 1.37|1.43|1.63                       \\
GT+\bootInv                         & 10.9|13.0|15.4               & 1.42|1.39|1.42                   & 0.00|3.29|4.07                    & 0.00|0.50|0.88                & 10.9|11.6|11.9               & 1.37|1.29|1.34                       \\ \hline
\bootInv                            & 9.4|11.8|14.3                & 5.01|4.06|3.56                   & 28.40|28.41|28.00                 & 2.59|2.78|2.91                & \textbf{10.9|11.6}|11.8      & 1.37|1.27|1.35                       \\ 
\rowcolor{my_gray}\supBoot           & \textbf{10.9}|13.0|15.4      & 1.95|1.79|1.62                   & 7.11|8.06|8.40                    & \textbf{0.64}|0.77|1.00       & \textbf{10.9|11.6|11.9}      & 1.37|1.34|1.40                       \\ \hline
\autoRF\cite{muller2022autorf}      & 3.6|7.9|10.6                 & 11.21|10.61|10.09                & 87.52|87.9|88.07                  & 6.04|5.99|5.95                & 10.0|9.4|8.8                 & 1.31|1.36|1.41                       \\ 
\autoRFWithFCOS                     & 7.5|15.0|17.2                & 1.34|0.87|0.81                   & 9.77|10.00|10.17                  & 0.85|0.81|0.78                & 9.8|10.5|10.5                & 1.29|1.19|1.16                       \\ 
\rowcolor{my_gray}\supNeRF           & 10.5|\textbf{16.4|18.8}      & \textbf{0.69|0.61|0.6}          & \textbf{7.01|7.01|7.07}           & 0.68|\textbf{0.70|0.73}        & 10.6|10.9|10.9               & \textbf{1.22|1.16|1.14}              \\

\end{tabular}
}
\end{footnotesize}
\end{table}

\begin{table}[ht]
\caption{\textbf{\kitti Cross-dataset Monocular Reconstruction and Pose Estimation Results}. We train all methods on \nuscenes dataset, and test on \kitti dataset. Our methods \supNeRF and \supBoot consistently show superior generalization in all metrics compared to the counterpart methods.
[Key: \textbf{Best}, FF = Feed Forward]
}
\label{tab:exp:main:kitti:ext}
\centering
\begin{footnotesize}
\scalebox{\scaleFraction}{
\begin{tabular}{l|c|c|c|c}
\hline
\multirow{2}{*}{Method}             & PSNR (\uparrowRHDSmall)                & DE (m) (\downarrowRHDSmall)         & RE (deg.) (\downarrowRHDSmall)    & TE (m) (\downarrowRHDSmall)   \\ \cline{2-5} 
                                    & FF|20it|50it                          & FF|20it|50it                 & FF|20it|50it               & FF|20it|50it             \\ \hline
GT (Freeze)+\bootInv                & 8.3|11.0|13.1                         & 1.33|1.09|0.82               & 0.00|0.00|0.00             & 0.00|0.00|0.00           \\
GT+\bootInv                         & 8.3|11.4|13.5                         & 1.33|1.10|1.22               & 0.00|3.24|4.02             & 0.00|0.48|0.89           \\ \hline
\bootInv                            & 7.0|10.0|12.3                         & 6.11|4.96|4.38               & 15.52|15.86|15.76          & 3.97|3.91|3.90           \\ 
\rowcolor{my_gray}\supBoot           & \textbf{7.6}|10.4|12.7                & 2.79|2.34|2,21               & 9.16|8.99|8.84             & 1.15|1.40|1.67           \\ \hline
\autoRF                             & 0.4|4.6|7.6                           & 9.89|9.28|8.83               & 89.67|90.18|90.42          & 6.16|6.11|6.07           \\ 
\autoRFWithFCOS                     & 2.4|11.2|13.4                         & 2.42|1.81|1.74                 & 11.95|12.48|12.5          & 2.2|2.14|2.09           \\
\rowcolor{my_gray}\supNeRF           & 4.0|\textbf{12.1|14.1}                & \textbf{2.19|1.6|1.54}        & \textbf{6.79|6.78|6.89}    & \textbf{1.06|1.04|1.01}  \\ \hline
\end{tabular}
}
\end{footnotesize}
\end{table}

\begin{table}[!tb]
\caption{\textbf{\waymo Cross-dataset Monocular Reconstruction and Pose Estimation Results}. We train all methods on \nuscenes dataset, and test on \waymo dataset.
[Key: \textbf{Best}, FF = Feed Forward]
}
\label{tab:exp:main:waymo:ext}
\centering
\begin{footnotesize}
\scalebox{\scaleFraction}{
\setlength\tabcolsep{0.15cm}
\begin{tabular}{lmc|c|c|c}
\myTopRule
\multirow{2}{*}{Method}     & PSNR (\uparrowRHDSmall)    & DE (m) (\downarrowRHDSmall)         & RE (deg.) (\downarrowRHDSmall)    & TE (m) (\downarrowRHDSmall)   \\ \cline{2-5} 
                            & FF|20it|50it               & FF|20it|50it                 & FF|20it|50it               & FF|20it|50it             \\ \myTopRule
GT(Frz)+\bootInv             & 9.2|10.4|11.4              & 3.16|2.94|2.44               & 0.00|0.00|0.00             & 0.00|0.00|0.00           \\
GT+\bootInv                  & 9.2|10.9|12.2              & 3.16|3.15|3.16               & 0.00|3.65|4.56             & 0.00|0.69|1.34           \\ \hline
\bootInv                     & 8.1|9.8|11.0               & 8.85|8.26|8.20               & 30.78|31.08|31.52          & 5.26|5.63|6.16           \\ 
\rowcolor{my_gray}\supBoot                & \textbf{8.6}|10.4|11.9     & 5.36|4.84|4.38               & 10.24|10.41|11.04          & 1.67|2.03|2.53           \\ \myTopRule
\autoRF                      & 0.6|6.6|9.8                & 6.76|6.71|6.53               & 86.56|87.39|87.67          & 9.1|9.11|9.14           \\ 
\autoRFWithFCOS               & 6.2|14.2|16.5             & 2.43|2.30|2.32               & \textbf{6.65|7.27|7.2}     & 3.22|3.25|3.31           \\
\rowcolor{my_gray}\supNeRF     & 4.8|\textbf{14.4|17.0}    & \textbf{2.32|1.67|1.56}      & 10.01|10.84|10.6           & \textbf{1.68|1.62|1.54}  \\ \hline
\end{tabular}
}
\end{footnotesize}
\end{table}

\section{Additional Ablation Study}
\label{sec:ablation:more}

We conducted a comprehensive ablation study to showcase the efficacy of our model design. 
First we presents an empirical analysis of \nerf-based optimization alone under varying initial errors, pose representations, as well two coordinate frames. 
After this, the study focuses on other possible alternatives of \supNeRF, analyzing the impact of freezing pose updates in NeRF phase, the impact of training with predicted \twoD boxes. Additionally, we explored the influence of pose refinement iterations and training epochs. It worth to mention that we count feed-forward iterations together with \nerf iterations together to make a total of 100 iterations for better visual comparison.





\subsection{Impact of initial error and pose representation to \nerf}
\label{sec:GBP:impact}

This ablation study focuses solely on the \nerf component to analyze its limitations under different initial errors and pose representations. From the results shown in \cref{fig:exp:vary:err:range}, five setups are evaluated on specific purpose: (i) To verify our claim on the coordinate frame choice, we conduct a \nerf pose optimization in C2O using an initial rotation error of $0.2$ radians. As expected, the rotation does not show any improvement. (ii) We perform the same experiment as in (i), but using the default O2C pose choice. The rotation exhibits some improvement, but not beyond $< 8^\circ$. (iii) We test the rotation error at $0.4$ radians, and while the rotation does improve, it appears to be unsatisfyingly slow, and the final error is still large when considering $8^\circ$ as the limit it could reach. (iv) Applying a longitudinal distance error with a ratio of $0.3$, where $T_{o2c}$ is randomly multiplied by a ratio of $0.7$ or $1.3$, does not effectively optimize the pose. This is due to the scale-depth ambiguity and difficulty in setting the \threeD translation step size at all distances. (v) We conduct the same test as in (iv) using relative translation, which yields a certain amount of improvement in the translation, but it is still not satisfying since the scale-depth ambiguity remains unaddressed. (vi) Finally, we combine a rotation error of $0.2$ radians and a translation error ratio of $0.3$, and optimize the relative pose space. This experiment also demonstrates very limited optimization of the pose.

Based on these observations, it can be concluded that direct application of the \nerf framework for pose optimization is very limited and may require additional modules or third-party methods to provide a good initial pose.

\begin{figure}[!t]
    \centering
    \includegraphics[width=0.99\textwidth]{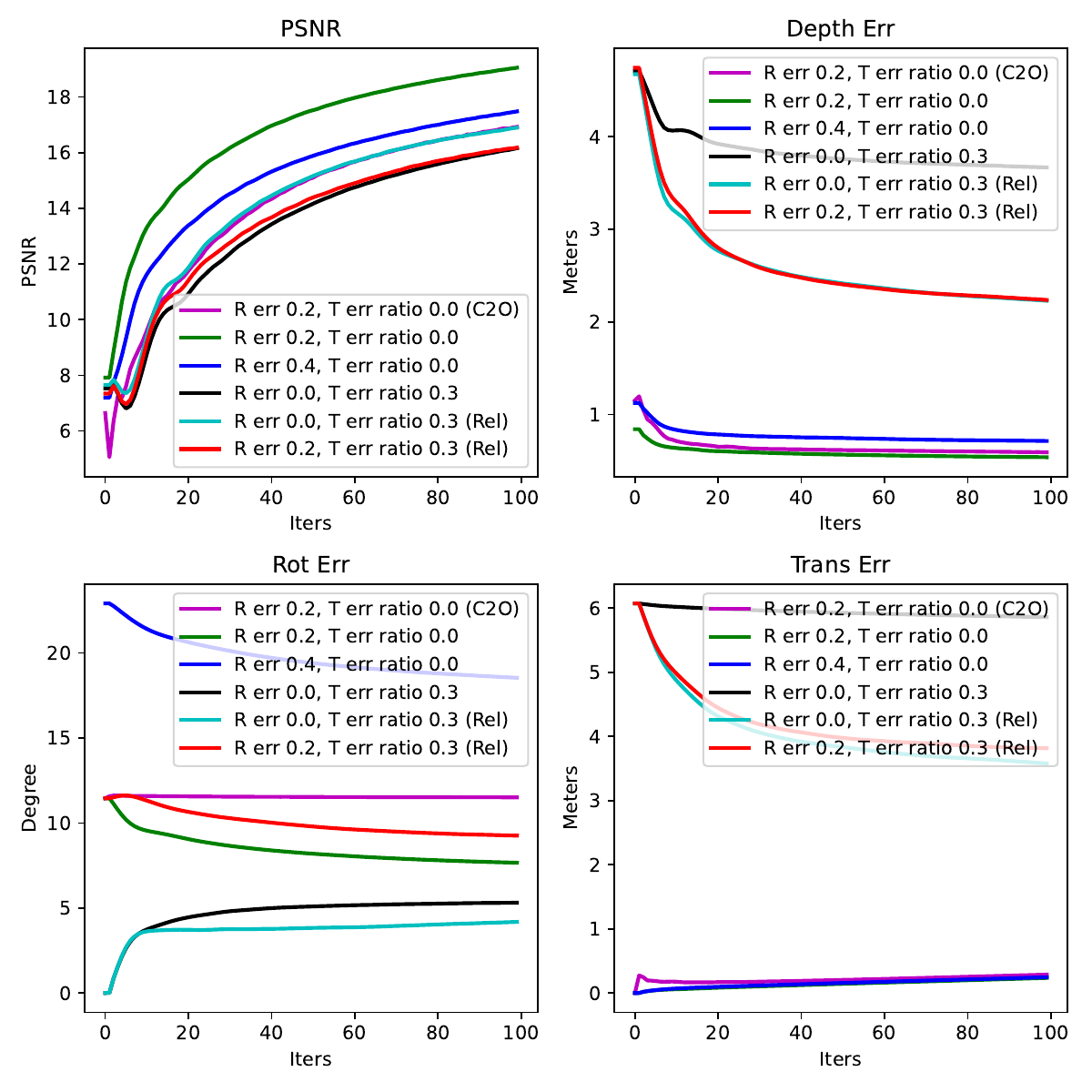}
    \caption{\textbf{Impact of Initial Pose Error and Pose Representation} to gradient-based \nerf pose optimization.}
  \label{fig:exp:vary:err:range}
\end{figure}

\subsection{Impact of freezing pose updating during NeRF}

In this study, we conducted a comparative analysis between \methodName and its alternative version, which lacks gradient-based pose updates, effectively keeping the feed-forward pose static throughout the NeRF process. This comparison was carried out across the validation splits of the \nuscenes, \kitti, and \waymo datasets to ensure a thorough evaluation. According to the results presented in \cref{tab:exp:ablation:freeze:pose}, omitting pose updates in the NeRF process generally results in a slight decline in PSNR values. However, the overall performance remains relatively similar across all evaluated metrics. Furthermore, we noticed that the integrated optimization of pose and shape within the NeRF framework sometimes causes deviations in the estimated pose from the actual pose, confirming our feed-forward pose estimation have reached a high accuracy beyond NeRF's capability to refine. By halting the pose update mechanism within NeRF, we can avert such decreases in pose accuracy.

\begin{table}[ht]
\caption{\textbf{Ablation on Freezing Poses in NeRF}. \supNeRF is evaluated on three datasets comparing to the version with frozen pose in NeRF phase. As observed, freezing NeRF pose leads to slightly lower PSNR but overall the two are very close in all metrics. Freezing the pose updates in NeRF can avoid potential degradation in pose accuracy in NeRF optimization stage. 
[Key: \firstKey{Best}, FF = Feed Forward, PF = Pose Freeze]
} 
\label{tab:exp:ablation:freeze:pose}
\centering
\begin{footnotesize}
\scalebox{0.72}{
\begin{tabular}{l|c|c|c|c|c|c}
\hline
\multirow{2}{*}{Method}             & PSNR (\uparrowRHDSmall)      & DE (m) (\downarrowRHDSmall)      & RE (deg.) (\downarrowRHDSmall)    & TE (m) (\downarrowRHDSmall)   & PSNR-C (\uparrowRHDSmall)    & DE-C (m) (\downarrowRHDSmall)        \\ \cline{2-7} 
                                    & FF|20it|50it                 & FF|20it|50it                     & FF|20it|50it                      & FF|20it|50it                  & FF|20it|50it                 & FF|20it|50it                         \\ \hline
\nuscenes                           & \textbf{10.5|16.4|18.8}      & \textbf{0.69|0.61|0.6}          & \textbf{7.01|7.01}|7.07            & \textbf{0.68}|0.70|0.73        & \textbf{10.6|10.9|10.9}     & \textbf{1.22|1.16|1.14}              \\
\nuscenes (PF)                      & \textbf{10.5}|16.1|18.3      & \textbf{0.69}|0.64|0.63          & \textbf{7.01|7.01|7.01}           & \textbf{0.68|0.68|0.68}        & \textbf{10.6}|10.8|10.8     & \textbf{1.22}|1.18|\textbf{1.14}     \\ \hline
\kitti                             & 4.0|\textbf{12.1|14.1}        & 2.19|1.6|\textbf{1.54}            & \textbf{6.79|6.78}|6.89           & \textbf{1.06|1.04|1.01}       & NA|NA|NA                     & NA|NA|NA                             \\ 
\kitti(PF)                         & \textbf{4.1}|12.0|13.7        & \textbf{2.15|1.57|1.54}           & 6.79|6.79|\textbf{6.79}           & \textbf{1.06}|1.06|1.06       & NA|NA|NA                     & NA|NA|NA                             \\ \hline
\waymo                             & \textbf{4.8|14.4|17.0}        & \textbf{2.32|1.67|1.56}           & \textbf{10.01}|10.84|10.6         & \textbf{1.68|1.62|1.54}       & NA|NA|NA                     & NA|NA|NA                             \\ 
\waymo(PF)                         & \textbf{4.8}|14.1|16.3        & \textbf{2.32}|1.74|1.72           & \textbf{10.01|10.01|10.01}        & \textbf{1.68}|1.68|1.68       & NA|NA|NA                     & NA|NA|NA                             \\ \hline

\end{tabular}
}
\end{footnotesize}
\end{table}

\subsection{Impact of training with predicted \twoD boxes}

To avoid any potential distractions, the primary experiments utilized ground-truth \twoD boxes. Therefore, we conducted an ablation study to investigate the impact of \twoD boxes as the first step. We utilized predicted \twoD boxes directly from the pre-trained Mask R-CNN~\cite{Detectron2018} for both training and testing. Additionally, we augmented the predicted \twoD boxes by randomly scaling them between 0.9 to 1.1 and introducing random jitters within 5 pixels during training to enhance the robustness of the unified model against noisy predicted \twoD boxes.

We evaluated the trained \supNeRF on the validation sets of both the \nuscenes dataset and \kitti dataset, as illustrated in \cref{fig:exp:impact:pred:box2d}. Our results indicated that using predicted \twoD boxes slightly lowered all scores for the \nuscenes test. However, it actually led to an improvement in the cross-dataset test on \kitti. The discrepancy in the definition of \twoD box ground-truth between the two datasets is the primary reason for this observation. \nuscenes uses a more lenient definition of a projected ground-truth box, whereas \kitti's definition is tightly annotated in \twoD. However, using the predicted \twoD boxes from the same Mask R-CNN removed this gap, resulting in a notable improvement in cross-dataset testing. 

\begin{figure}[!t]
    \centering
    \includegraphics[width=0.9\textwidth]{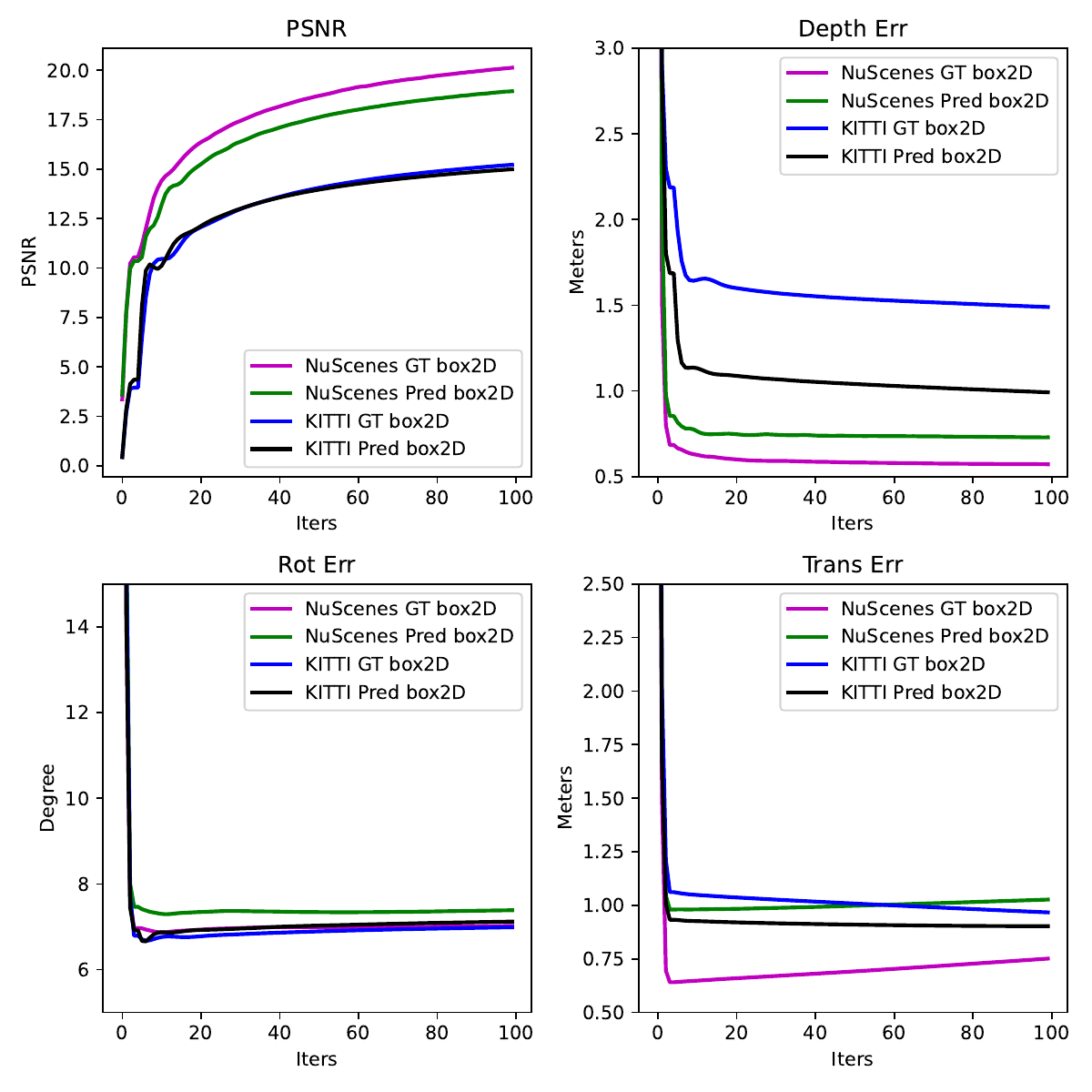}
    \caption{\textbf{Impact of Training with Predicted \twoD Boxes on \supNeRF.} Although training with predicted 2D boxes leads to worse in-domain performance, it in contrary improve the generalization in cross-domain test. Note that the degradation of in-domain test of translation estimation from later NeRF gradient-based updates indicate the feed-forward steps have reached a high pose accuracy beyond NeRF's capability to refine.}
  \label{fig:exp:impact:pred:box2d}
\end{figure}




\subsection{Impact of pose refine iterations}

We conducted an analysis of the performance of \supNeRF on the \nuscenes dataset, comparing the effectiveness of different numbers of iterations for the pose estimation module. As depicted in \cref{fig:exp:vary:dl:iter}, our findings suggest that performance reaches its peak at iteration 3. This could be attributed to the optimal number of iterations used in the training process. Moreover, as we observed that \nerf optimization did not yield any further improvements in rotation or translation beyond iteration 3, it is likely that this iteration count is sufficient for \nerf to perform accurate neural reconstruction

\begin{figure}[!t]
    \centering
    \includegraphics[width=0.8\textwidth]{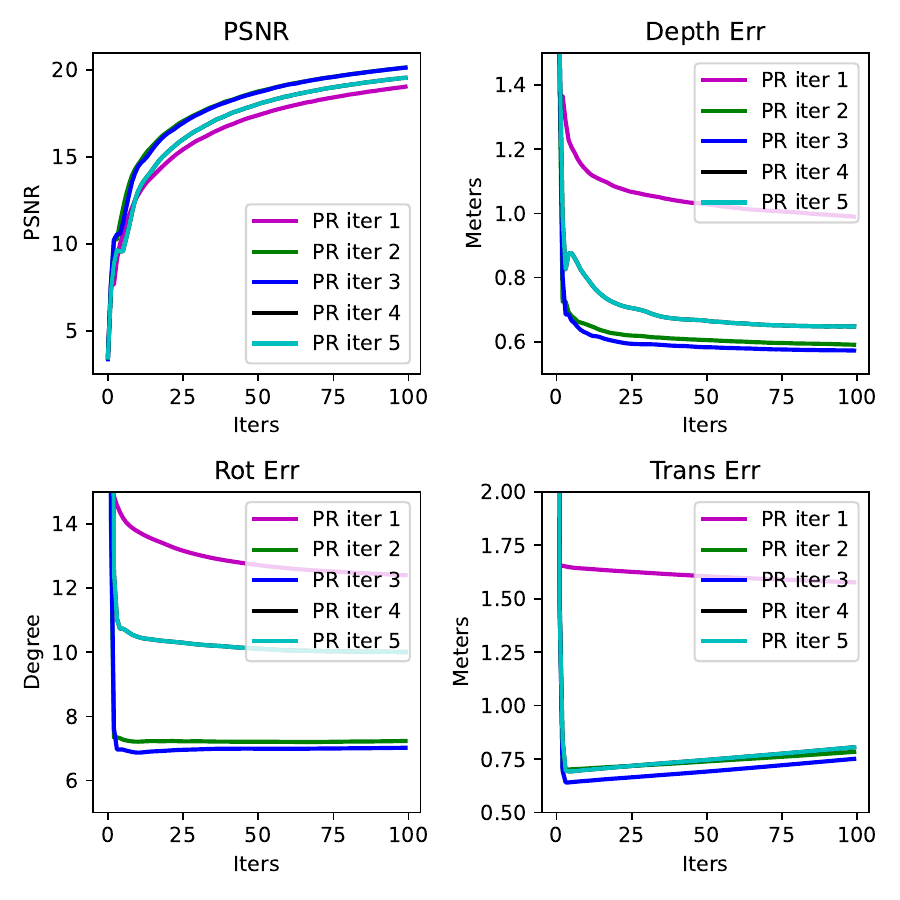}
    \caption{\textbf{Impact of Number of Feed-forward Iterations in Pose Refinement.} We observe using 3-iterations of our feed-forward pose refinement leads to the optimal performance. Note that the pose estimation performance achieves its peak at the end of feed-forward iterations. The degradation of in-domain test of translation estimation from later NeRF gradient-based updates indicate the feed-forward steps have reached a high pose accuracy beyond NeRF's capability to refine.}
  \label{fig:exp:vary:dl:iter}
\end{figure}

\subsection{Impact of training epochs}

We conducted a hyper-parameter study to determine the optimal number of training epochs for our model on the \nuscenes dataset. As shown in \cref{fig:exp:vary:epoch}, the performance of our model improves with the number of epochs, and it reaches the peak at Epoch 40. However, we also observed a slight decline in translation performance beyond Epoch 30. Therefore, we decided not to use more epochs to avoid overfitting to the training dataset.

\begin{figure}[!t]
    \centering
    \includegraphics[width=0.8\textwidth]{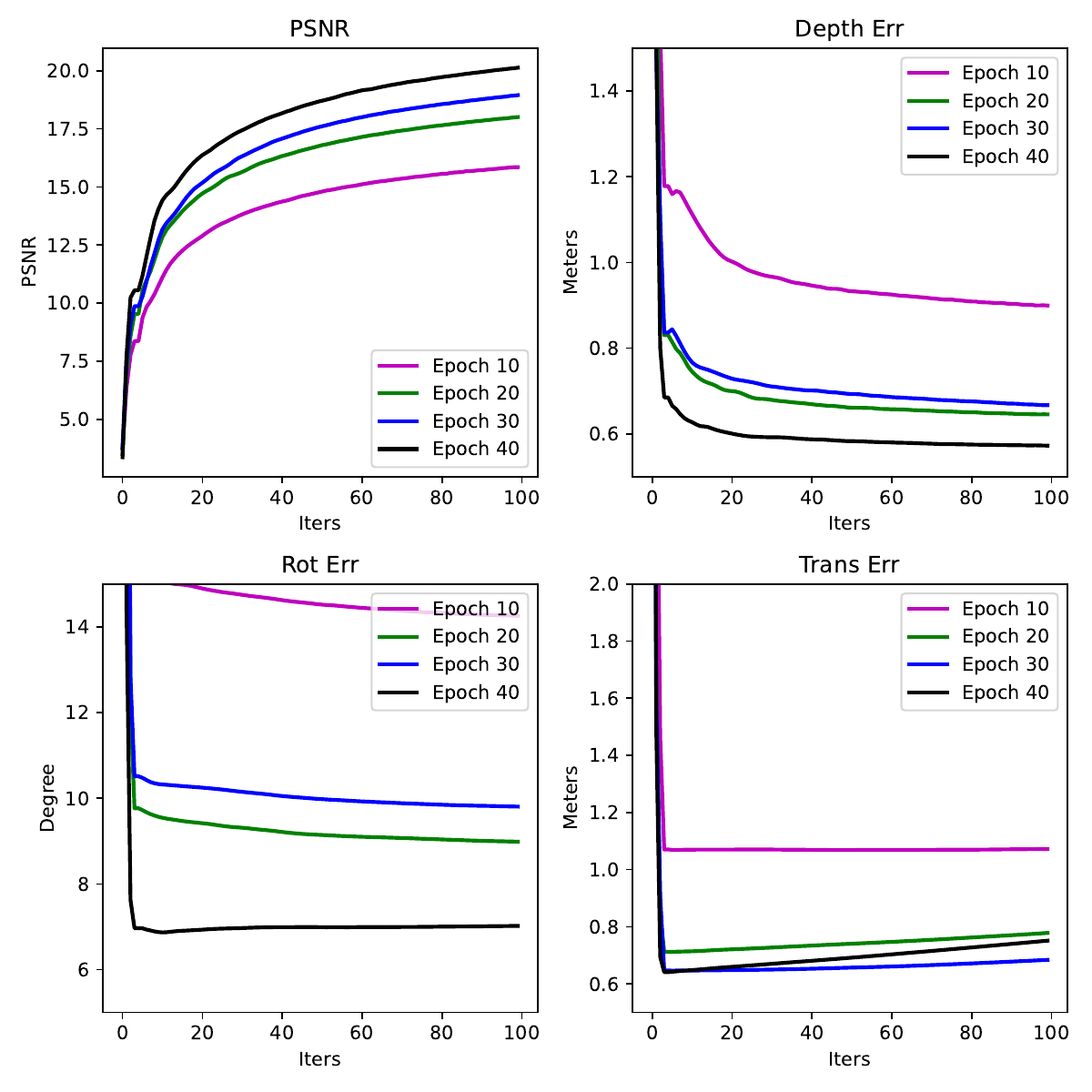}
    \caption{\textbf{Impact of Training Epochs.} We observe the more epoch \methodName got trained, the better in-domain performance it got. We did not go beyond 40 epochs to prevent overfitting to the training dataset. All the evaluation results are produced with models trained for epochs on \nuscenes. The degradation of in-domain test of translation estimation from later NeRF gradient-based updates indicate the feed-forward steps have reached a high pose accuracy beyond NeRF's capability to refine.}
  \label{fig:exp:vary:epoch}
\end{figure}


\section{Dataset Preparation}
\label{sec:data:prepare}

In the data preparation process, we first filtered for sequences captured during the daytime and utilized a pre-trained Mask R-CNN~\cite{Detectron2018} to obtain instance masks since \nuscenes does not provide \twoD segmentation masks. We then matched the provided \threeD bounding box annotations with the resulting instance masks and categorized the instance masks into foreground, background, and unknown regions, following the same occlusion mask preparation process as \autoRF. We excluded severe image truncation cases and applied additional rules for object selection. Specifically, we only considered objects having \twoD regions of interest (ROI) no less than 2500 pixels, an intersection-over-union (IOU) of no less than 0.5 between its mask outbox and the \twoD box projected from \threeD box, and within 40 meters distance. To ensure challenging and diverse test samples, we also used 5 Lidar points to select testing samples for \nuscenes and 10 points for \kitti, removing the 10\% at the bottom of the \threeD object box before counting. Our selection rule is expected to include more distant and occluded samples compared to \autoRF~\cite{muller2022autorf} as we focus more on auto-labeling and joint optimization of pose and \nerf.

\begin{figure}[!t]
    \centering
    \includegraphics[width=0.4\textwidth]{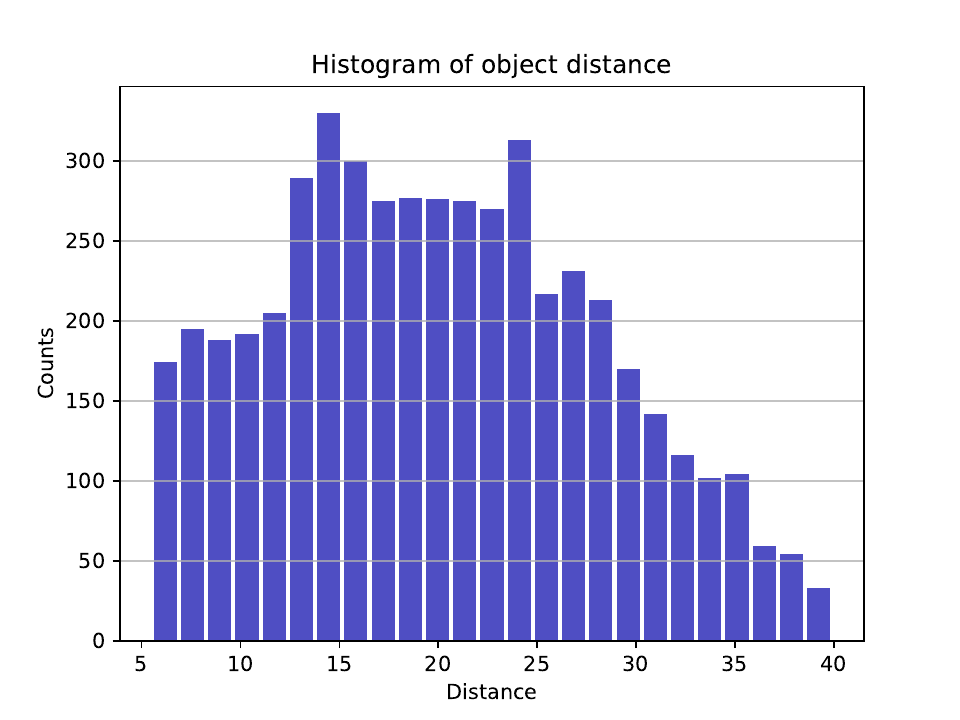}
    \includegraphics[width=0.4\textwidth]{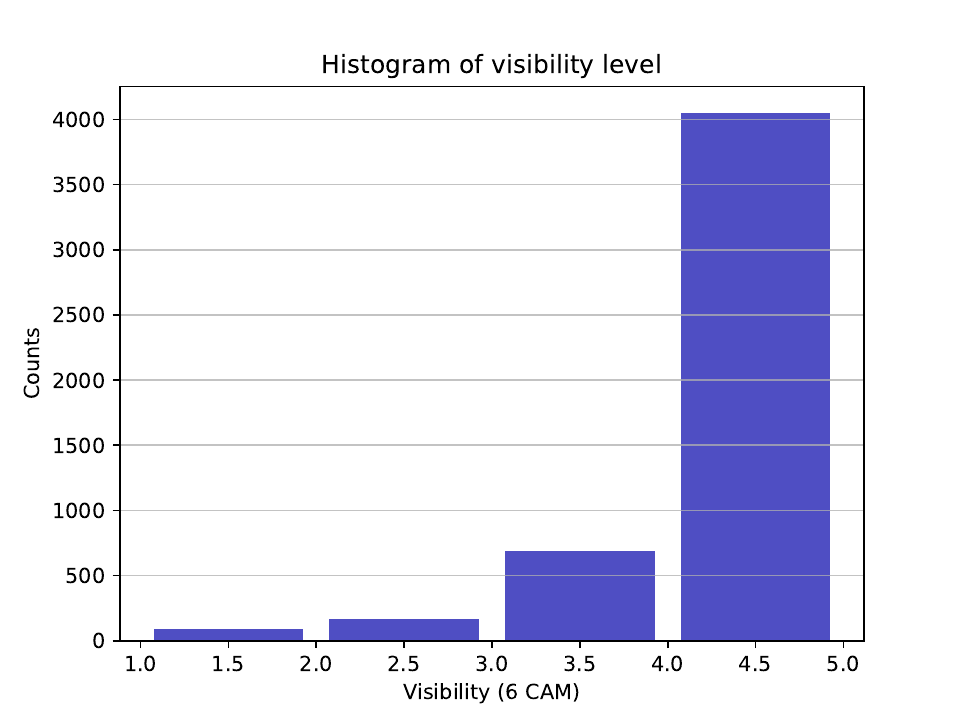}
    \caption{Data Distribution of the Curated Subset for \nuscenes.}
  \label{fig:nusc_hist}
\end{figure}

\begin{figure}[!t]
    \centering
    \includegraphics[width=0.4\textwidth]{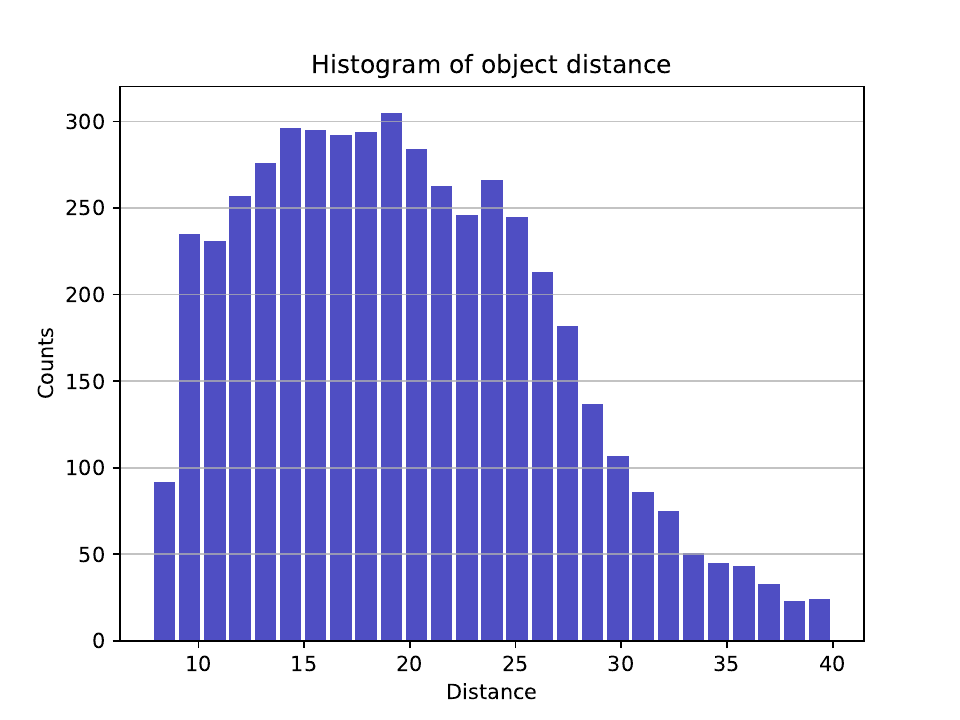}
    \includegraphics[width=0.4\textwidth]{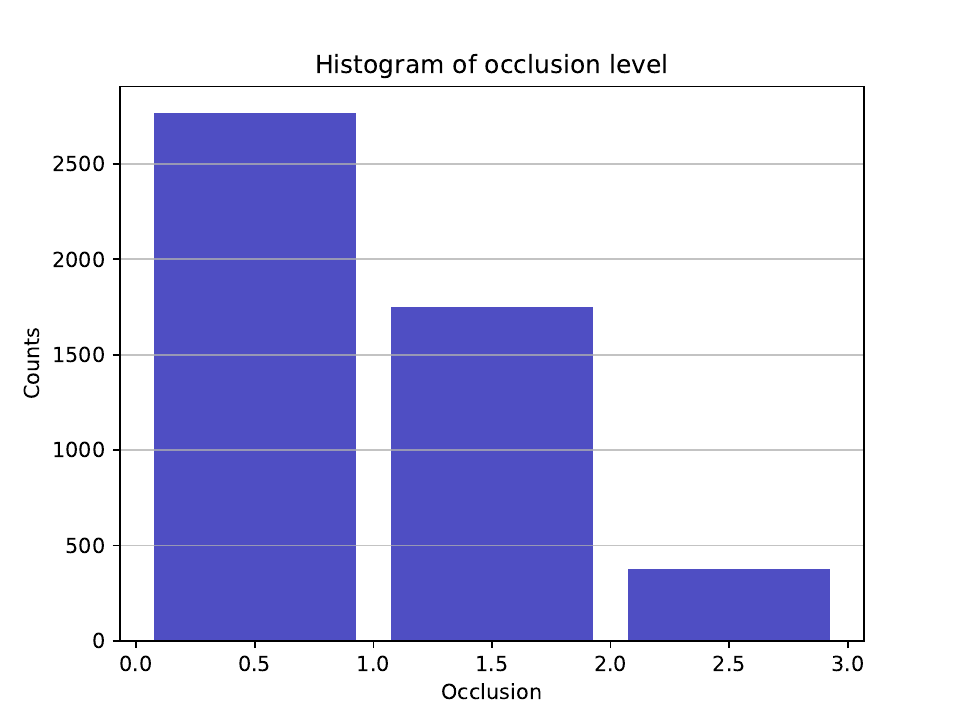}
    \caption{Data Distribution of Curated Subset for \kitti.}
  \label{fig:kitti_hist}
\end{figure}

To better understand the difficulty of the datasets used in this study, we plotted histograms of the object distance and level of visibility for the \nuscenes dataset in \cref{fig:nusc_hist}, and histograms of the object distance and level of occlusion for the \kitti dataset in \cref{fig:kitti_hist}. For the \nuscenes dataset, we used a pre-defined visibility scale where 1 indicates 0-40\% visibility, 2 indicates 40-60\% visibility, 3 indicates 60-80\% visibility, and 4 indicates 80-100\% visibility across six cameras. In contrast, for the \kitti dataset, we defined occlusion levels as 0 for fully visible, 1 for partly occluded, 2 for largely occluded, and 3 for unknown. The object distance distributions of the two datasets are similar, but \nuscenes has more visible objects overall. It is important to note that the \nuscenes visibility scale applies to all 6 cameras, so single-camera visibility may be much lower, but this information is not available.


\section{Detailed Running Speed Analysis}
\label{sec:detail:running:speed}

Notably, the efficiency of image encoding depends on the resolution of image input, which we set to $128 \times 128$ for all experiments. \nerf's efficiency, on the other hand, depends on the resolution of the rendered image patch during online optimization as well as the total number of iterations. 
To ensure efficiency for \nerf iterations, we use a patch resolution of $32 \times 32$ in online optimization. 
It's also worth noting that the last rendering stage can employ any resolution due to the implicit representation of \nerf, and we do not include its cost in our overall computational analysis. This set of analysis is all conducted on a single Nvidia A5000 graphic card for fair comparison.

As we mentioned the speed limitation of NeRF can been effectively tackled by recent advances in neural rendering~\cite{MullerESK_2022_SIGGRAPH_instantNGP,Kerbl:etal:SIGGRAPH2023:GaussianSplatting}, which we consider as orthogonal to our contribution to the feed-forward stage. However, in hypothetical scenarios where we substitute the original \nerf with the more efficient INGP~\cite{MullerESK_2022_SIGGRAPH_instantNGP}, the \nerf process could speed up by up to 100 times. \cref{tab:run:time:ext} shows our adapted framework SUP-INGP could then reach real-time performance with 50 iterations, surpassing the \fcosThreeD-INGP pipeline significantly.

\begin{table}[!tb]
\caption{\textbf{Model Size and Running Time Comparison} in feed-forward scenarios, and with 20 and 50 iterations of \nerf optimization.
\supNeRF gets the smallest running time.
[Key: \firstKey{Best}, FF= Feed Forward]
}
\label{tab:run:time:ext}
\centering
\scalebox{\scaleFraction}{
\begin{tabular}{l m c | c | c | c }
\myTopRule
Method                                  & Params (M) (\downarrowRHDSmall)           & FF (s)   (\downarrowRHDSmall)   & FF+20it (s)  (\downarrowRHDSmall)      & FF+50it (s)    (\downarrowRHDSmall)       \\
\myTopRule
\multirow{2}{*}{\autoRFWithFCOS~\cite{muller2022autorf}}                            & 91.116                                    & 0.123                           & 0.714                                  & 1.599          \\
                                        & (36.166+54.950)                           & (0.114+0.009)                   & (0.114+0.600)                          & (0.114+1.485)       \\
\rowcolor{my_gray}\textbf{\supNeRF}      & \first{49.816}                            & \first{0.018}                   & \first{0.608}                          & \first{1.493}           \\ \hline
\bootInv~\cite{Pavllo_2023_CVPR}        & 182.616                                   & 0.156                           & 3.534                                  & 8.601           \\ 
\rowcolor{my_gray}\textbf{\supBoot}      & \first{57.580}                            & \first{0.018}                   & \first{3.396}                          & \first{8.463}           \\ 
\hline
\fcosThreeD+INGP(Virt)                  & $\sim$100                                 & 0.123                           & $\sim$0.129                            & $\sim$0.138     \\
\rowcolor{my_gray}SUP-INGP(Virt)          & $\sim$\first{50}                          & \first{0.018}                   & $\sim$\first{0.024}                    & $\sim$\first{0.033}     \\
\myTopRule
\end{tabular}
}
\end{table}

Our proposed unified method represents a minimum viable extension to the existing \nerf pipeline. The majority of the computation is dedicated to the iterative \nerf fitting to image pixels, while the image encoder (ResNet50) and the 3-iteration pose estimator (implemented using a few MLP layers upon the latent codes) are highly efficient. In particular, the average running time for the image encoding is $0.009s$, the 3-iteration pose estimation takes $0.009s$, and the remaining 50 iterations of \nerf take $1.475s$, resulting in a total of $1.493s$ per image patch. If to only count the feed-forward stage, it takes only $0.018s$ per image patch. When we estimate the running time of the virtual pipeline SUP-INGP(Virt), we keep the feed-forward time the same at $0.018s$, but consider the \nerf optimization 100 times faster, which leads to $1.475|100 = 0.015s$. Summing up the two leads to $0.033s$ as presented in \cref{tab:run:time}, which could be considered a real-time pipeline. A similar way of calculation was done for \fcosThreeD+INGP(Virt) as well.


\begin{figure*}[!t]
    \centering
    \includegraphics[width=0.99\textwidth]{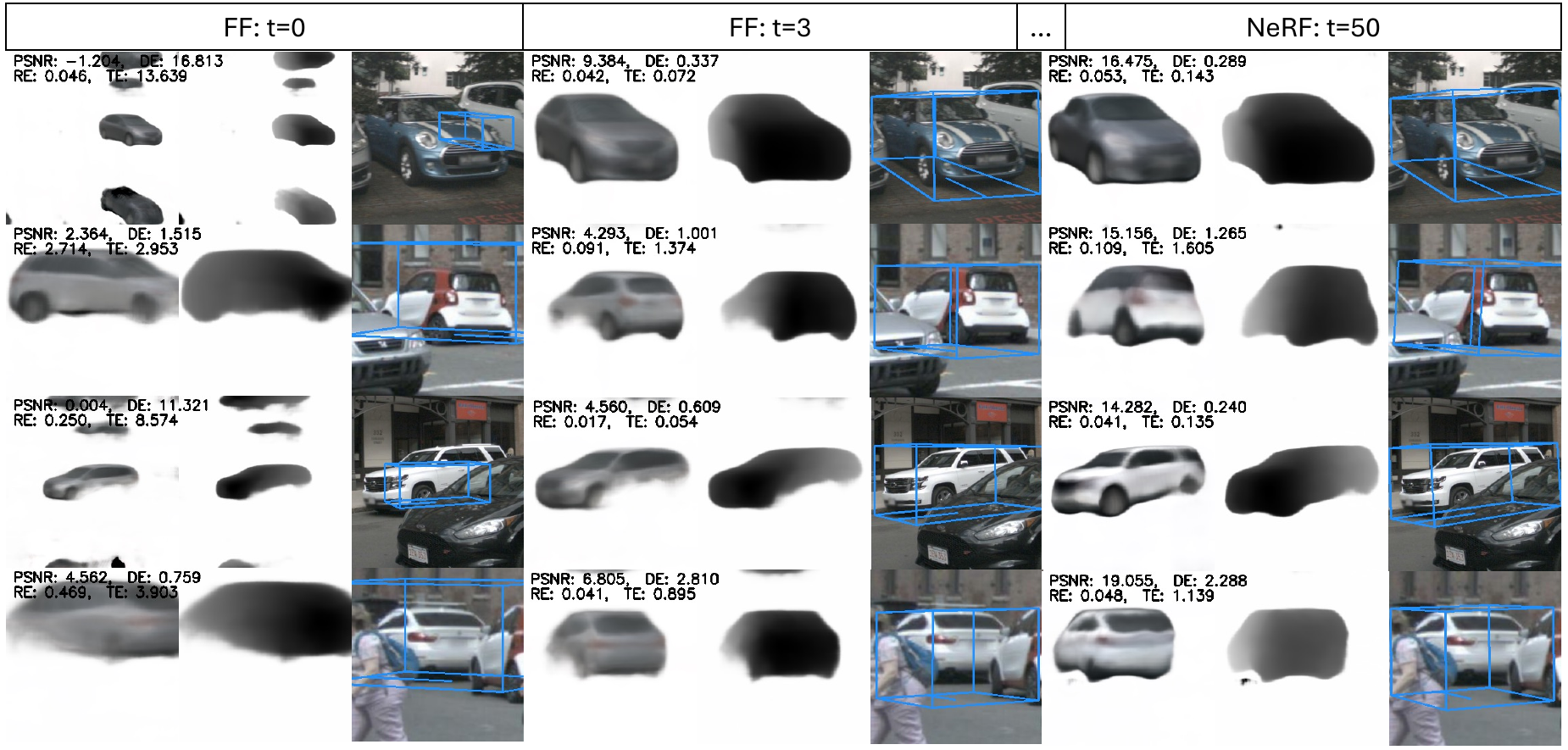}
    \includegraphics[width=0.99\textwidth]{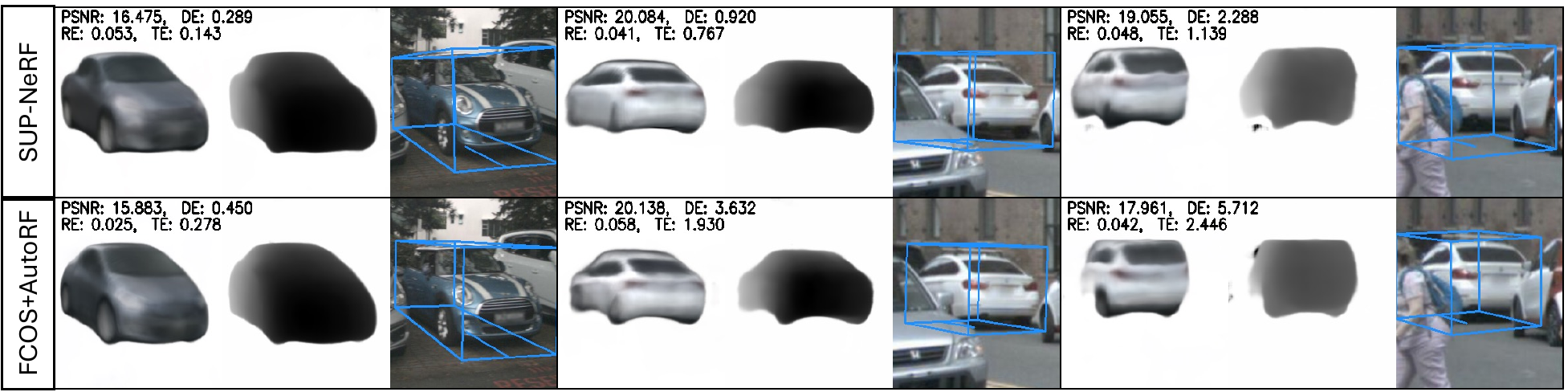}
    \caption{\textbf{Qualitative Results on \nuscenes Dataset.} In the top panel, we demonstrate \supNeRF executes pose estimation reliably, fast converging from a random initial pose to the true one, and enables neural reconstruction under diverse object poses, occlusion cases under this cross-dataset setup. In the bottom panel, \supNeRF is visually compared to the \fcosThreeD+\autoRF, demonstrates sharper rendered image, higher accuracy in shape and pose.}
  \label{fig:exp:visual:results:nusc:supp}
\end{figure*}

\begin{figure*}[!t]
    \centering
    \includegraphics[width=0.99\textwidth]{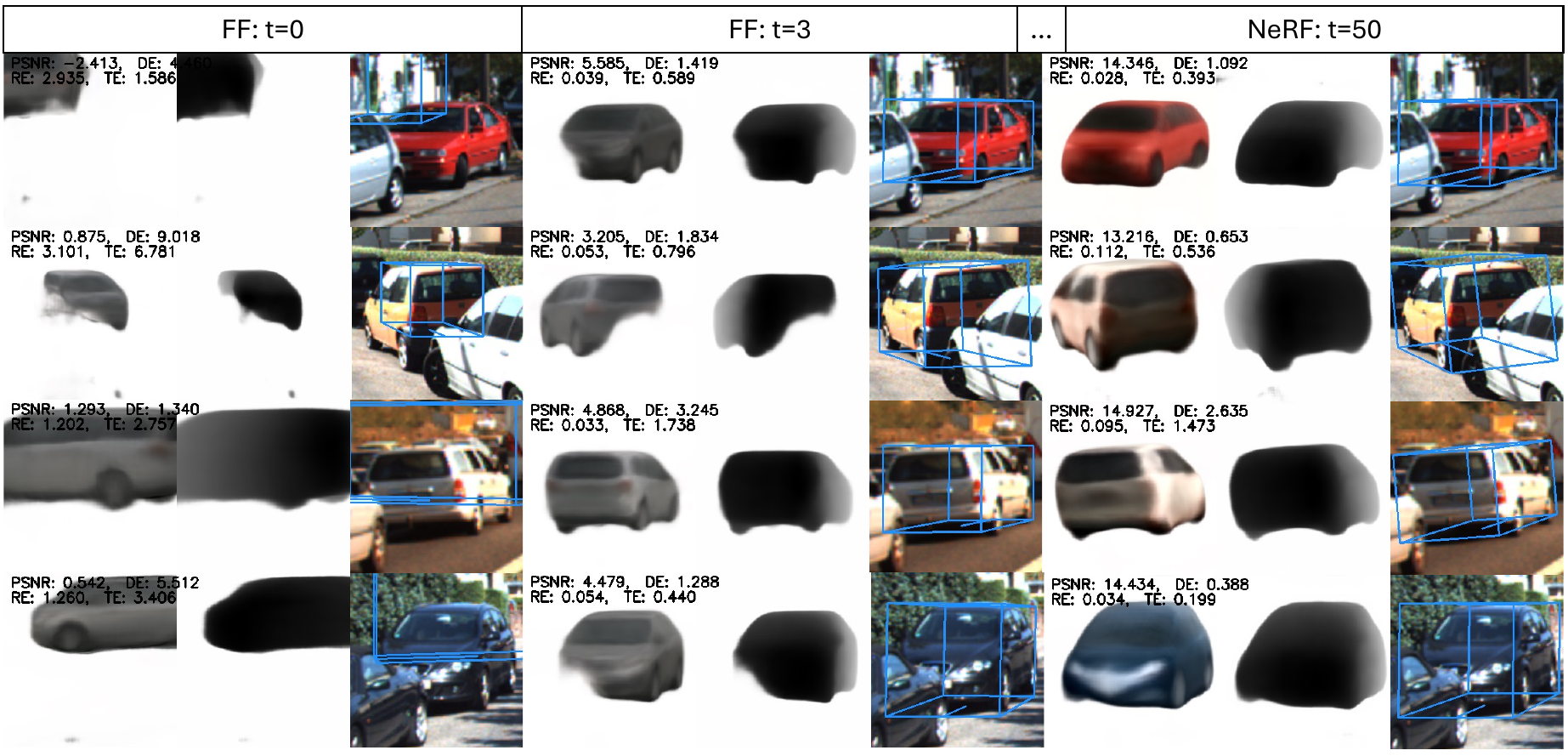}
    \includegraphics[width=0.99\textwidth]{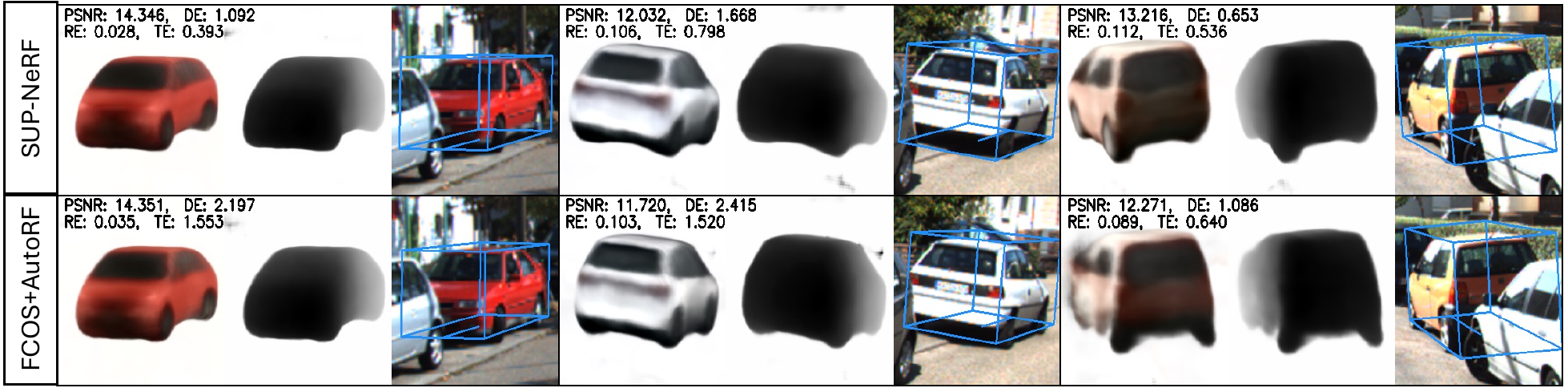}
    \caption{\textbf{Qualitative Results on \kitti Dataset.} The upper half includes visualization of \methodName's iterative process. The lower half includes visual comparison to \fcosThreeD+\autoRF on three examples.}
  \label{fig:exp:visual:results:kitti:supp}
\end{figure*}

\section{Additional Visual Results}
\label{sec:more:vis}

\subsection{Visual comparison of \autoRF-based pipelines}

We present additional visual results of our \autoRF-based pipeline, \supNeRF, on both \nuscenes and \kitti datasets, comparing to \autoRF + \fcosThreeD. More over, we also visualize the iterative progress of our joint estimation from feed-forward pose estimation stage to \nerf optimization stage. For visualization purposes, we normalize the depth image to better reflect the shape rather than the actual depth.

As observed in the top panels in \cref{fig:exp:visual:results:nusc:supp,fig:exp:visual:results:kitti:supp}, \supNeRF effectively estimates the right pose even under significant occlusion. The iterative \nerf phase shows promising completion of occluded shapes to some extent. Although \supNeRF appears superior to the other two in completing shapes (probably due to better pose), we also observe that none of these \autoRF-based monocular frameworks can always guarantee the success in completing the object shape from such challenging monocular setup.

More specifically, comparing our unified model to the combination of separately trained models \autoRF + \fcosThreeD, we observe that \supNeRF performs better, particularly in \nerf encoding and decoding. This indicates that joint training of pose benefits the \nerf more than the pose estimation. In contrast, the improvement on the pose estimation side is less significant. When compared to \autoRF + \fcosThreeD, our approach shows a significant improvement in pose estimation in the Cross-dataset test on \kitti. Although the same \autoRF is used, the lower performance of \fcosThreeD in pose estimation leads to weaker \nerf reconstruction end result. 

These visual results also highlight some limitations of the current approach. First, occlusion handling still remains to be a major challenge, particularly in completing missing shape information rather than over-fitting to the visible portion. Secondly, domain gap is also observed to be active challenge: while \supNeRF shows some Cross-dataset generalization from the pose regression aspect, the performance of monocular reconstruction in a Cross-dataset setup showed a notable downgrade.

\subsection{Visual comparison of \bootInv-based pipelines}

In this section, we showcase additional visual results comparing \bootInv and our \supBoot pipeline on the \nuscenes dataset in \cref{fig:exp:bootinv:nusc}, and on the \kitti dataset in \cref{fig:exp:bootinv:kitti}. We selected examples with varying degrees of occlusion to demonstrate the resilience of both methods in handling such challenges. The results are visually represented through single renderings of complete object images, depth maps and normal maps. Furthermore, since \nuscenes provides views of objects from different angles, we randomly selected another view for rendering the object images, depth maps, normal maps, based on the given pose. The novel-view rendering results also reflect the cross-view evaluation presented in \cref{tab:exp:main:nusc:ext}.  However, as the \kitti dataset does not have multi-view observations of the same object, we randomly rotated the reconstructed objects to a different orientation only for visualization purpose.

Across both datasets, we observed that the original \bootInv struggles with anything beyond minimal occlusion, while \supBoot shows significantly better robustness in pose estimation under occlusion. We also noted that inaccurate pose estimation greatly impacts both shape and appearance reconstruction. The primary reason for \bootInv's poor pose estimation performance with occluded objects is due to its reliance on NOCS + PnP for initial pose estimation. We have observed that NOCS predictions become noisy in occluded cases, leading to significant errors in PnP pose estimation, both in terms of rotation and translation. However, from a neural reconstruction perspective, the pre-trained decoder from \bootInv (integrated into our pipeline with frozen weights) demonstrates a strong ability to maintain object completeness and capture accurate color themes, even with incorrect poses and severe occlusions. This aligns with its superior cross-view performance compared to \autoRF-based pipelines. Additionally, we noted limitations in the shape and texture space of the decoder, particularly evident in the last example from the \nuscenes visualization, \cref{fig:exp:bootinv:nusc}. When the observed object's shape and texture fall outside the pre-trained model's distribution, the reconstructed shape and appearance can deviate significantly from the actual object. This highlights the ongoing research challenge of balancing prior knowledge with precise current observation recovery, especially in occluded situations.

\begin{figure*}[!t]
    \centering
    \includegraphics[width=0.99\textwidth]{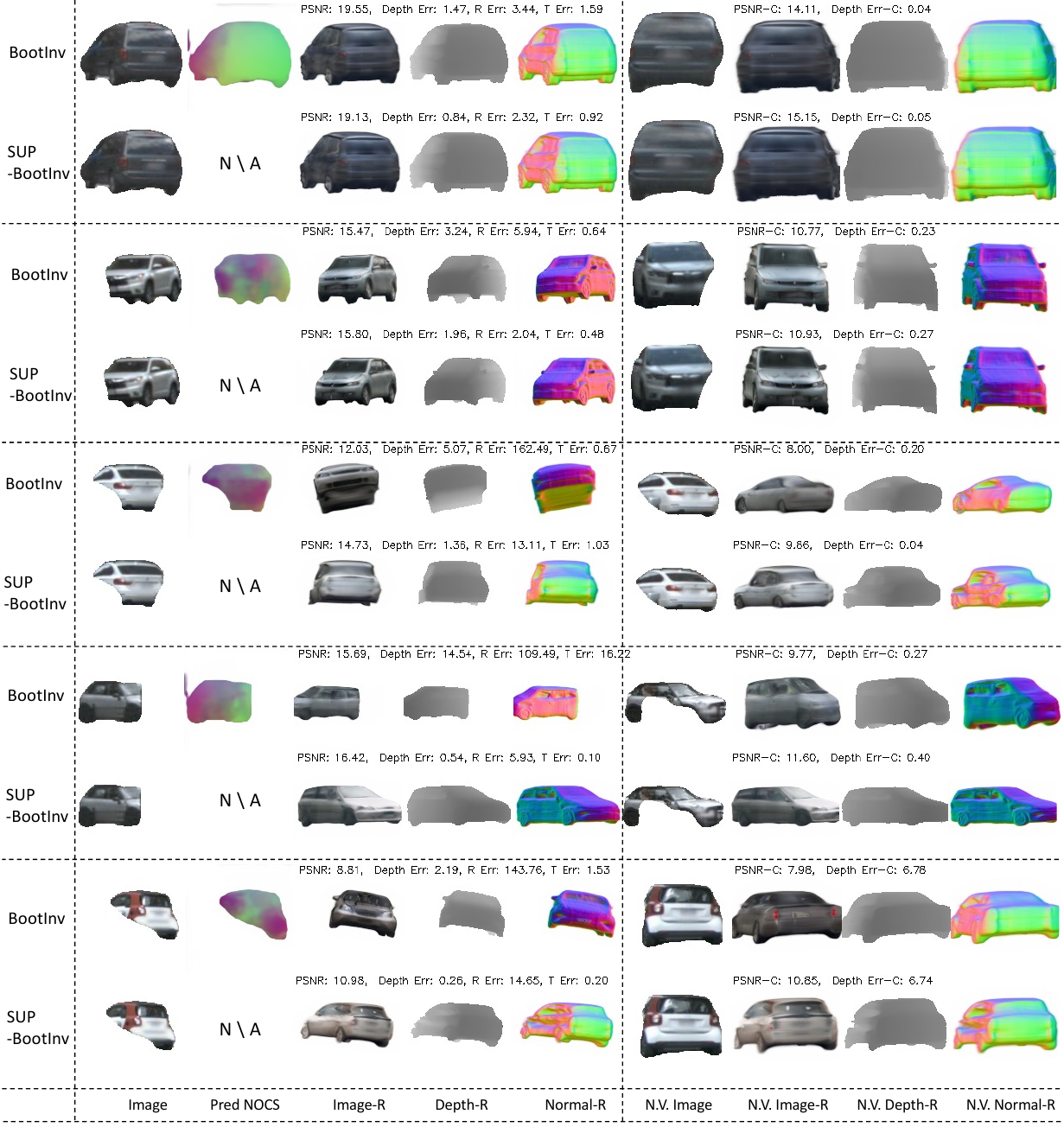}
    \caption{\textbf{Visual Comparison of \bootInv and \supBoot on \nuscenes Dataset.} The Novel View (N.V.) Rendering (-R) is based on the object pose from a real image capturing the same object from another view.}
  \label{fig:exp:bootinv:nusc}
\end{figure*}

\begin{figure*}[!t]
    \centering
    \includegraphics[width=0.99\textwidth]{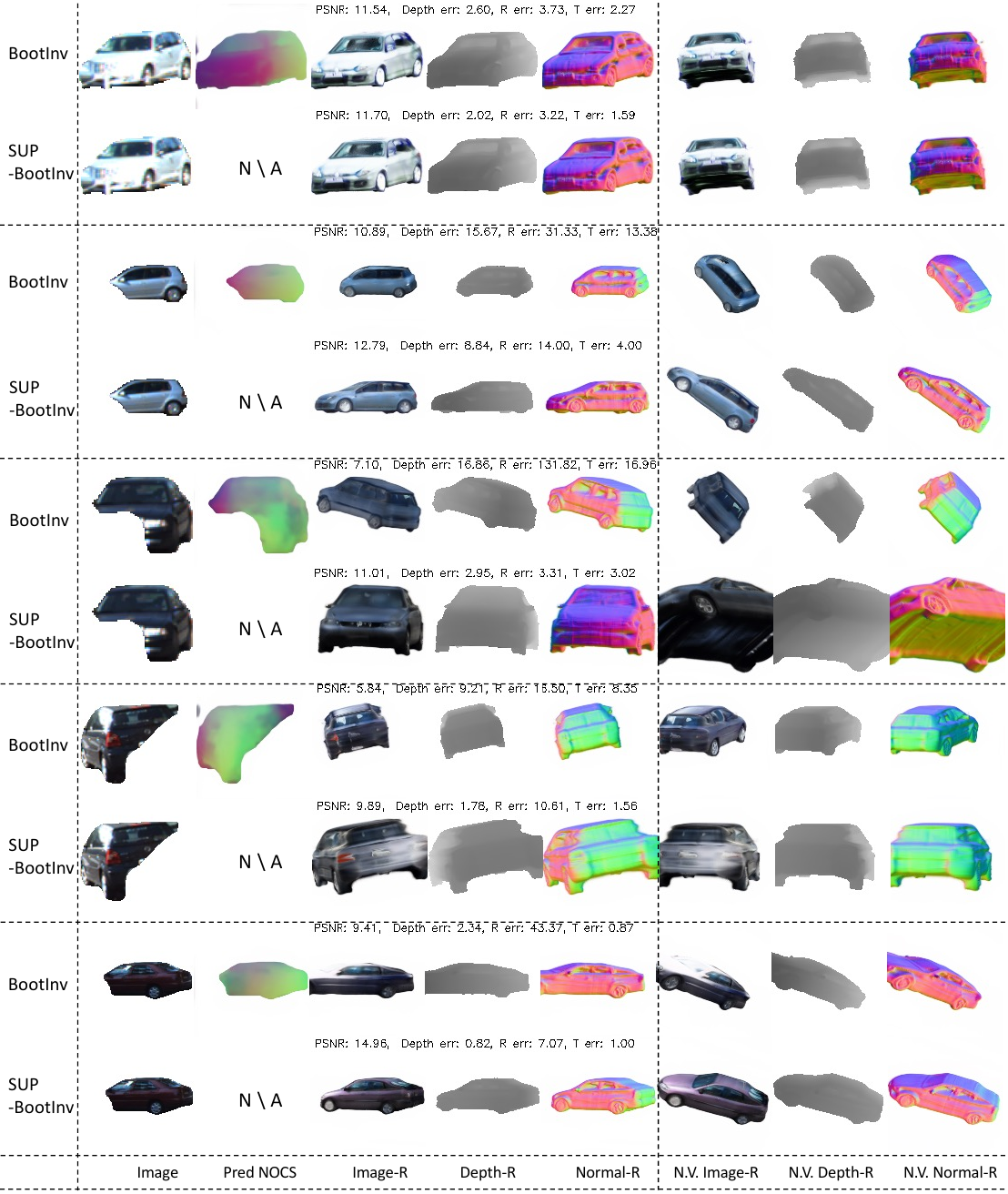}
    \caption{\textbf{Visual Comparison of \bootInv and \supBoot on \kitti Dataset.} The Novel View (N.V.) Rendering (-R) is based on random pose so that the two methods may not be directly compared on the same second view.}
  \label{fig:exp:bootinv:kitti}
\end{figure*}

\subsection{Visualization on cross-view evaluation}

To better illustrate the cross-view evaluation procedure, we provide additional visual examples in \cref{fig:exp:cross:eval:4rows} which is based on our model \supNeRF. Here, each row showcases different views of a unique object, with the first sample in each row serving as the basis for optimization and the remaining ones used for cross-view assessment. In our practical experiments, we treat each view of an object as the optimizing view in turn, creating a matrix-like array of sample sets for comprehensive evaluation, as depicted in \cref{fig:exp:cross:eval:full:pairs}.

The visual outcomes reveal that the optimization quality is influenced by the complexity of the optimizing image. Challenges such as inaccuracies in masks and severe occlusions can significantly impair reconstruction quality and pose estimation accuracy. Moreover, factors like varying illumination, along with asymmetric reconstruction, are identified as additional contributors to suboptimal results. These observations highlight pivotal areas for future research, particularly in contexts demanding certain efficiency standards.

\begin{figure*}[!t]
    \centering
    \includegraphics[width=0.99\textwidth]{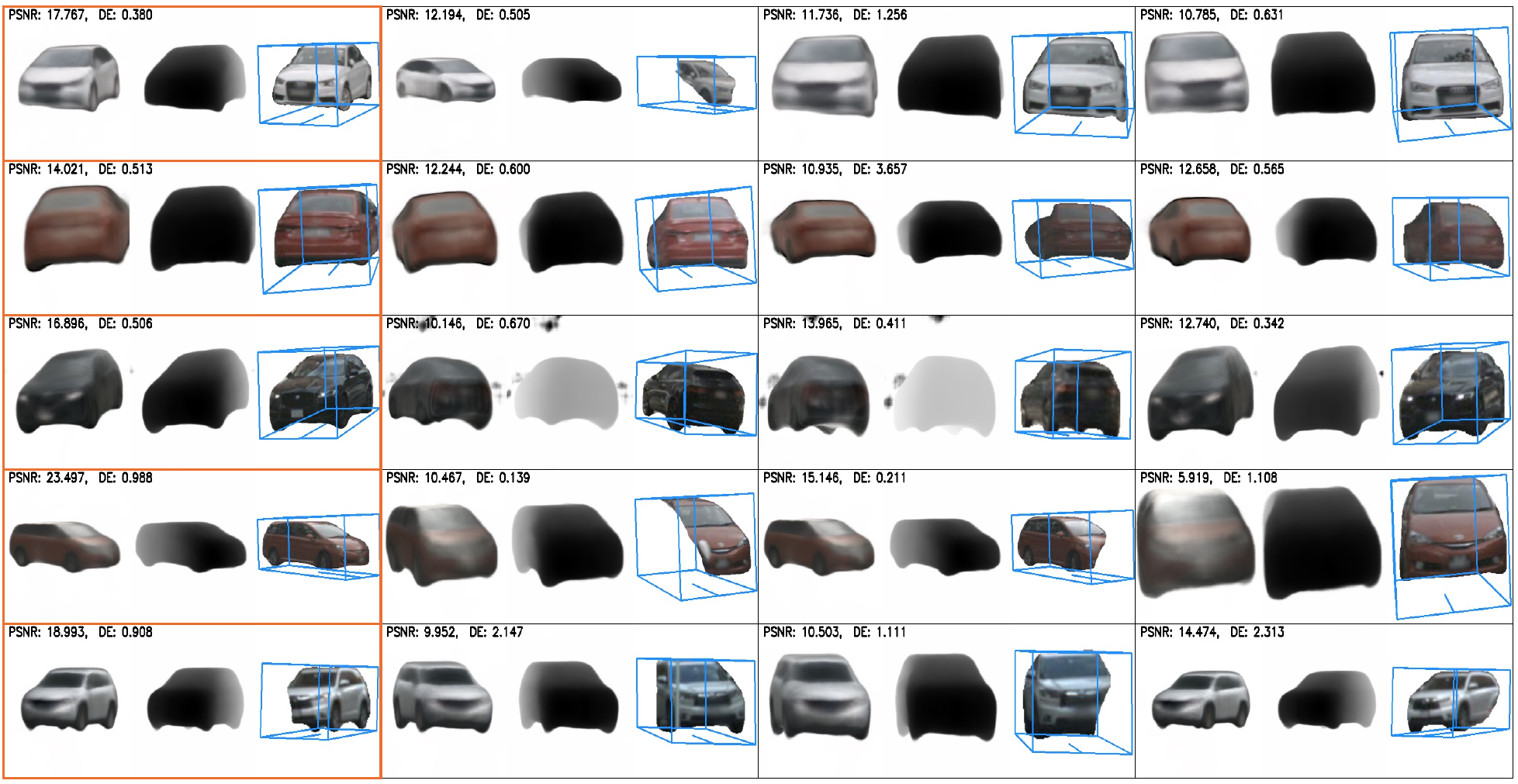}
    \caption{\textbf{Visualization of the Cross-view Evaluation.} Each row presents a series of image samples capturing the same object from various perspectives. The orange-highlighted sample serve as the basis for monocular pose estimation and \nerf reconstruction, while the remaining samples in each row exclusively employ the reconstructed shape and texture codes for rendering and PSNR/depth error assessment. For each sample, we display the rendered image, rendered depth map, and the original image with an object mask and box, in that order. Notably, the boxes shown on the optimization images are based on the predicted poses, whereas in the novel views, we use ground-truth poses.}
  \label{fig:exp:cross:eval:4rows}
\end{figure*}

\begin{figure*}[!t]
    \centering
    \includegraphics[width=0.99\textwidth]{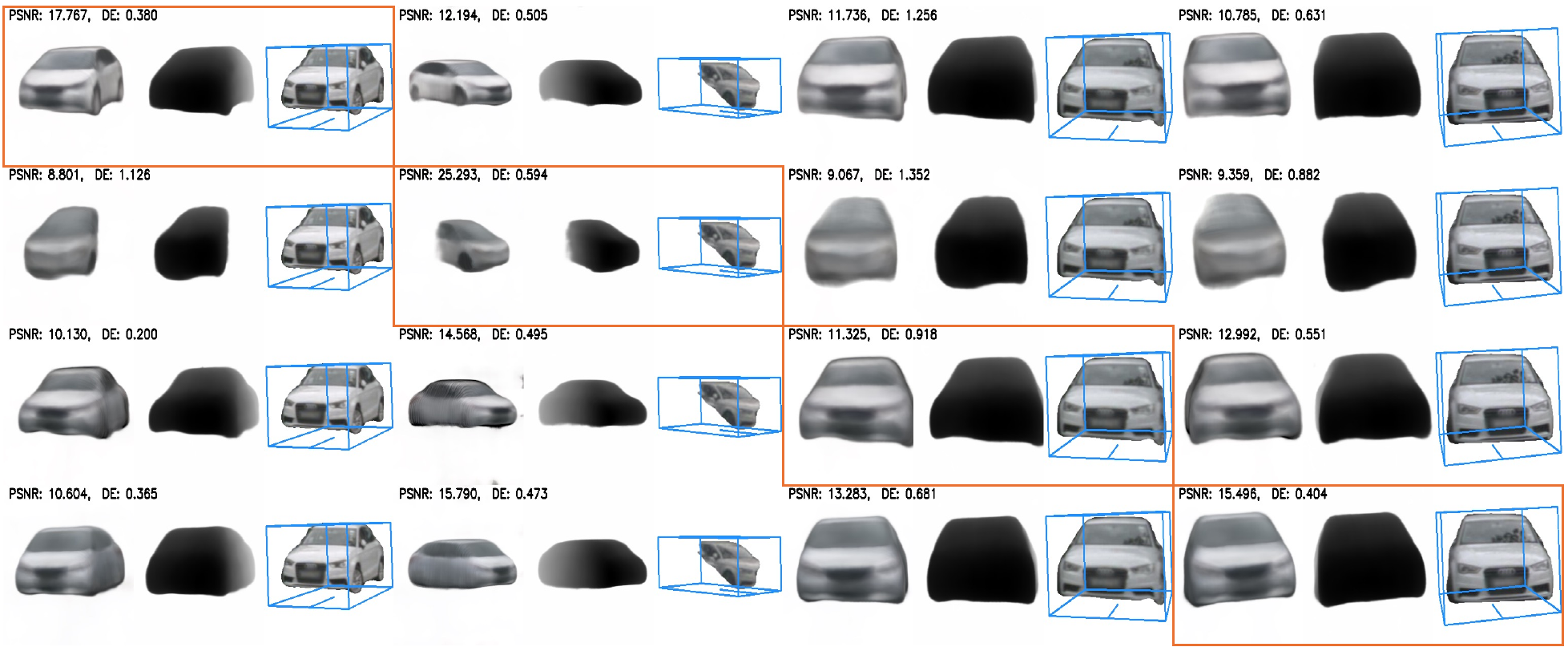}
    \caption{\textbf{Visualization of All Cross-View Pairs of One Object in Evaluation.} For a set of multi-view images, each sample take turns to be the optimization image as shown in each row. The orange-highlighted example serves as the basis for monocular pose estimation and \nerf reconstruction, while the remaining samples in each row exclusively employ the reconstructed shape and texture codes for rendering and PSNR/depth error assessment. The cross-view evaluation is based on ground-truth poses and focuses on the monocular reconstruction quality.}
  \label{fig:exp:cross:eval:full:pairs}
\end{figure*}
\end{document}